\documentclass[runningheads]{llncs}

\usepackage{eccv}

\usepackage{eccvabbrv}

\usepackage{graphicx}
\usepackage{wrapfig}
\usepackage{multirow}
\usepackage{booktabs} 
\usepackage{caption}
\usepackage{pifont}
\usepackage{bbding}
\usepackage[accsupp]{axessibility}  

\usepackage{hyperref}

\usepackage{orcidlink}

\newcommand{\norm}[2]{\left \lVert #1 \right \rVert_#2}

\newcommand{\expect}[2]{\mathbb{E}_{#2}\Biggl[#1\Biggr]}

\newcommand*\samethanks[1][\value{footnote}]{\footnotemark[#1]}

\begin{document}

\title{FRDiff : Feature Reuse for Universal Training-free Acceleration of Diffusion Models} 

\titlerunning{FRDiff}

\author{Junhyuk So\thanks{Equal Contribution}\inst{1}\orcidlink{0000-0002-9210-1284} \and
Jungwon Lee\samethanks\inst{2}\orcidlink{0009-0006-6985-2916} \and
Eunhyeok Park\inst{1,2}\orcidlink{0000-0002-7331-9819}}

\authorrunning{J. So et al.}

\institute{Department of Computer Science and Engineering, \and Graduate School of Artificial Intelligence, \\
POSTECH, Pohang, South Korea \\
\email{\{junhyukso,leejungwon,eh.park\}@postech.ac.kr}}

\maketitle

\begin{abstract}
The substantial computational costs of diffusion models, especially due to the repeated denoising steps necessary for high-quality image generation, present a major obstacle to their widespread adoption. While several studies have attempted to address this issue by reducing the number of score function evaluations (NFE) using advanced ODE solvers without fine-tuning, the decreased number of denoising iterations misses the opportunity to update fine details, resulting in noticeable quality degradation. In our work, we introduce an advanced acceleration technique that leverages the temporal redundancy inherent in diffusion models. Reusing feature maps with high temporal similarity opens up a new opportunity to save computation resources without compromising output quality. To realize the practical benefits of this intuition, we conduct an extensive analysis and propose a novel method, FRDiff. FRDiff is designed to harness the advantages of both reduced NFE and feature reuse, achieving a Pareto frontier that balances fidelity and latency trade-offs in various generative tasks.
\end{abstract}

\begin{figure}[h!]
    \centering
        \includegraphics[width=0.99\columnwidth]{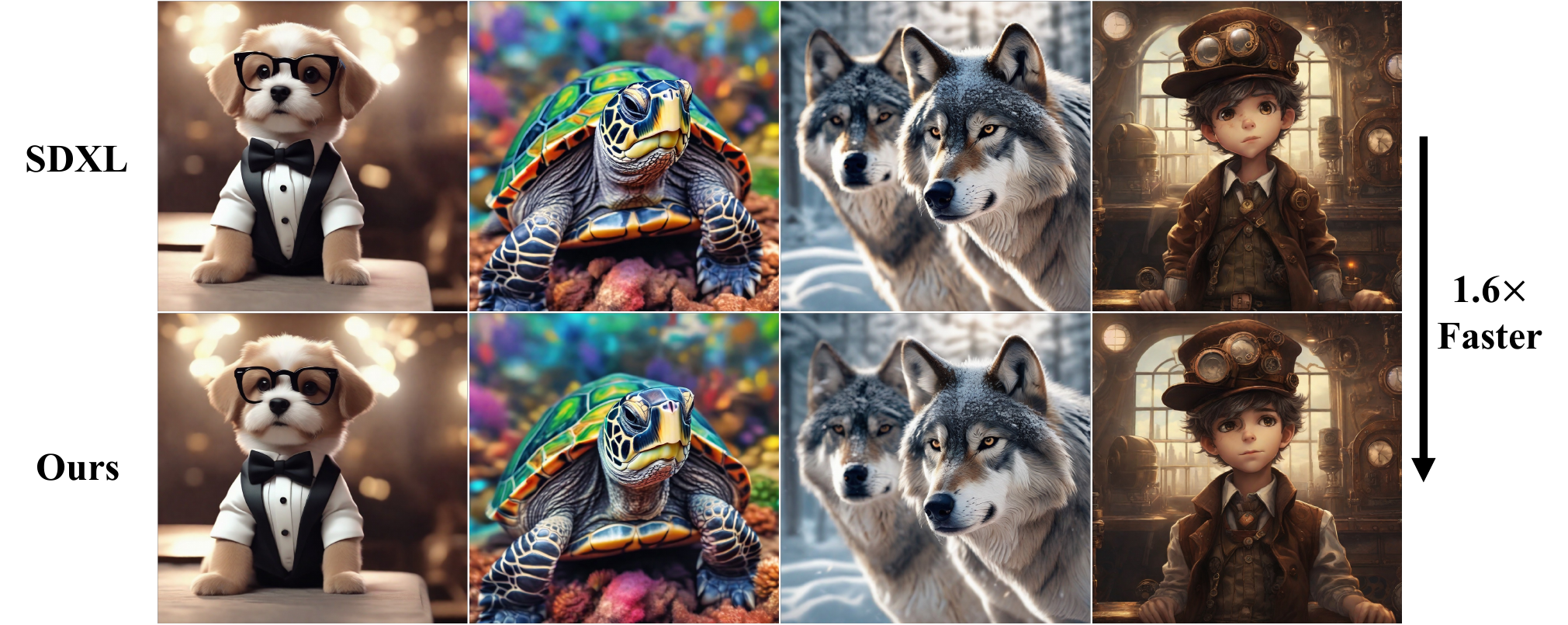}
    \caption{SDXL Acceleration with FRDiff}
    \label{fig:main_fig}
\end{figure}

\section{Introduction}
\label{sec:intro}

The diffusion model has gained attention for its high-quality and diverse image generation capabilities~\cite{dalle,ldm,imagen, sdxl}. Its outstanding quality and versatility unlocked new potentials across various applications, including image restoration~\cite{stablesr,diffbir}, image editing~\cite{ip2p,dds,pnp,mdp,preditor}, conditional image synthesis~\cite{universal-guidance,mcm-diffusion,lawdiffusion,controlnet,boxdiff,freedom,t2i-adapter}, and more. However, the substantial computation cost of the diffusion model, particularly due to its dozens to hundreds of denoising steps, poses a significant obstacle to its widespread adoption. To fully harness the benefits of diffusion models in practice, this performance drawback must be addressed.

Recently, many studies have proposed methods to mitigate the computational burden of diffusion models. A representative approach involves a zero-shot sampling method \cite{ddim,pndm,song2020score}, which typically employs advanced ODE or SDE solvers capable of maintaining quality with a reduced number of score function evaluations (NFE). While these methods demonstrate the potential for acceleration without fine-tuning, the performance improvement achievable within the accuracy margin is often insufficient. On the other hand, another direction employs learning-based sampling methods \cite{salimans2022progressive, meng2023distillation, song2023consistency, liu2022flow}, applying fine-tuning to preserve generation quality with a reduced NFE. However, the requirement of fine-tuning, such as additional resources and a complex training pipeline, makes it challenging to use in practice. \textit{To realize performance benefits in practice with minimal constraints, we need more advanced zero-shot methods with higher potential}.

In this work, we focus on an important but overlooked aspect of the diffusion models. Since they entail iterative denoising operations, \textbf{the feature maps within the diffusion models exhibit temporal redundancy}. According to our extensive analysis, specific modules within diffusion models show considerable similarity in their feature maps across adjacent frames. By reusing these intermediate feature maps with higher temporal similarity, we can significantly reduce computation overhead while maintaining output quality. Building on this insight, we propose a new optimization potential named \textbf{feature reuse (FR)}. However, the naive use of FR doesn't guarantee superior performance compared to the conventional reduced NFE method. Our thorough experiments reveal that FR has distinctive characteristics compared to reduced NFE methods, and both methods can complement each other to maximize the benefits we can achieve. 

Overall, \textbf{we propose a comprehensive method named FRDiff, designed to harness the strengths of both the reduced NFE and FR}. Specifically, we introduce a score mixing technique to generate high-quality output with fine details. Additionally, we design a simple auto-tuning, named \textbf{Auto-FR}, to optimize the hyperparameters of FR to maximize the outcome quality within given constraints, such as latency. This approach can be applied to any diffusion model without the need for fine-tuning in existing frameworks with minimal modification. We conduct extensive experiments to validate the effectiveness of FRDiff on various tasks in a zero-shot manner. We can achieve up to a \textbf{1.76x} acceleration without compromising output quality across a range of tasks, including a task-agnostic pretrained model for text-to-image generation, as well as task-specific fine-tuned models for super resolution and image inpainting. Code is available at \url{https://github.com/ECoLab-POSTECH/FRDiff}.

\section{Related Works}

\subsection{Diffusion Models}

The diffusion model, introduced in \cite{sohl2015deep}, defines the forward diffusion process by gradually adding Gaussian noise at each time step. Conversely, the reverse process generates a clean image from random noise by gradually removing noise from the data.
In DDPM \cite{ho2020denoising}, the authors simplified the diffusion process using a noise prediction network $\epsilon_\theta(x_t,t)$ and reparameterized the complex ELBO loss \cite{kingma2013auto} into a more straightforward noise matching loss. 
On a different note, \cite{song2020score} transforms the forward process of the diffusion model into a Stochastic Differential Equation (SDE). 
More recently, Classifier-Free Guidance (CFG) \cite{ho2022classifier} has been introduced to guide the score toward a specific condition $c$. In the CFG sampling process, the score is represented as a linear combination of unconditional and conditional scores.

Since FRDiff is formulated based on the temporal redundancy inherent in the iterative diffusion process, it can be seamlessly integrated into all the previously mentioned methods, providing benefits irrespective of their specific details.

\subsection{Diffusion Model Optimization}

To accelerate the generation of diffusion models, many studies have concentrated on reducing NFE, which can be broadly categorized into two groups: zero-shot sampling \cite{ddim,gddim,pndm,edm,lu2022dpm,xu2023restart,lu2022dpmv2}, applying optimization to the pre-trained model, and learning-based sampling \cite{salimans2022progressive,meng2023distillation,latentconsistency,liu2022flow}, involving an additional fine-tuning.

Zero-shot sampling methods typically employ advanced Ordinary Differential Equation (ODE) solvers capable of maintaining generation quality even with a reduced NFE. For instance, DDIM \cite{ddim} successfully reduced NFE by extending the original DDPM to a non-Markovian setting and eliminating the stochastic process. 
Furthermore, methods utilizing Pseudo Numerical methods \cite{pndm}, Second-order methods \cite{edm}, and Semi-Linear structures \cite{lu2022dpm,lu2022dpmv2} have been proposed to achieve better performance. Learning-based sampling finetunes the model to perform effectively with reduced NFE. For example, Progressive Distillation \cite{salimans2022progressive} distills a student model to achieve the same performance with half the NFE. Recently, the consistency model \cite{song2023consistency,latentconsistency} successfully reduced NFE to 1-4 by predicting the trajectory of the ODE.

In addition, there are studies aimed at optimizing the backbone architecture of the diffusion model. These studies involve proposing new diffusion model structures \cite{ldm,bksdm}, as well as lightweighting the model's operations through techniques such as pruning \cite{diffprune}, quantization \cite{qdiff, tdq}, and attention acceleration \cite{tome}.

In this work, we primarily focus on enhancing the benefit of zero-shot model optimization for diffusion models. However, it's important to note that the proposed method could be applied in conjunction with other learning-based or backbone optimization studies.

\begin{figure}[t]
    \centering
        \includegraphics[width=0.97\columnwidth]{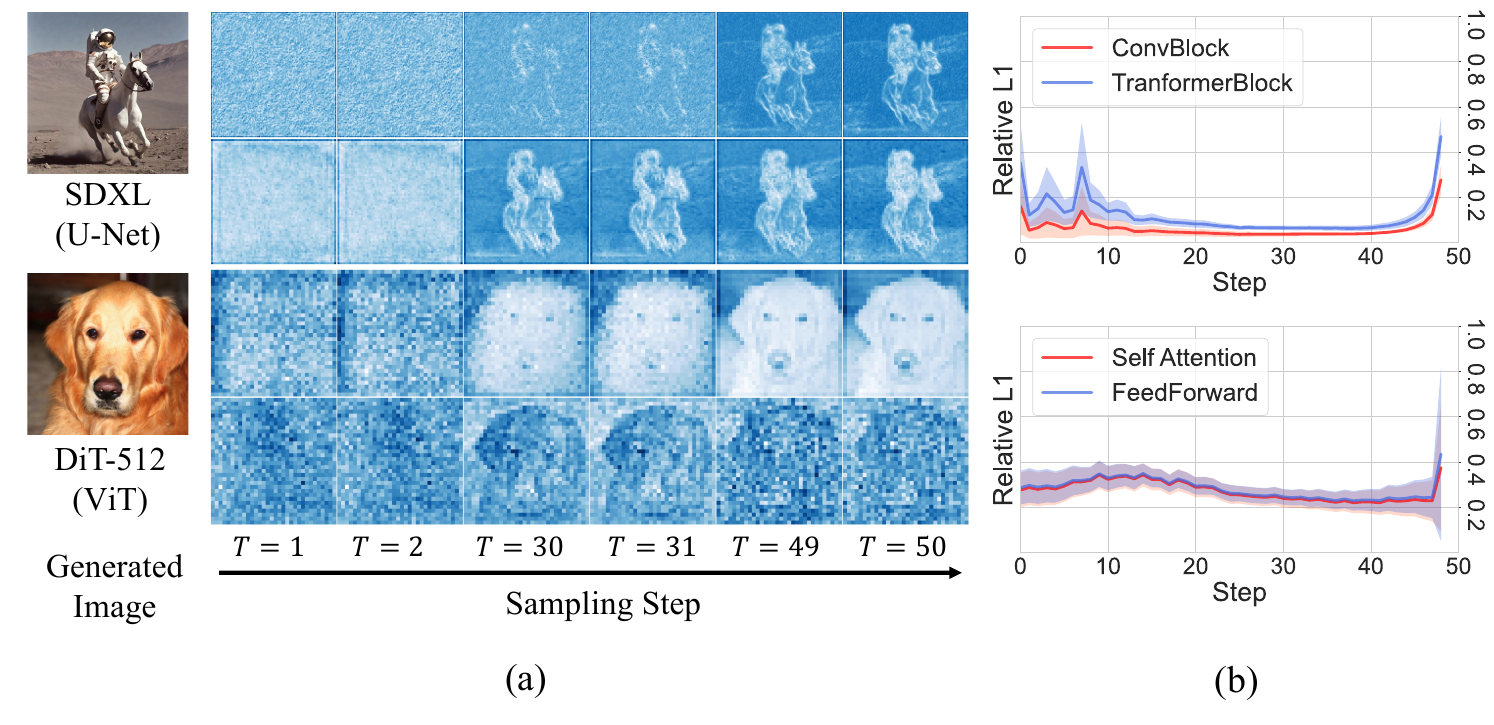}
    \caption{Temporal Similarity Analysis of Diffusion Model: (a) Visualization of Intermediate Feature Maps During Inference. (b) Mean and Variance of Relative L1 Distance Between Adjacent Time Steps.}
    \label{fig:temporal_similarity}
\end{figure}

\section{Method}
In this study, we introduce the idea of feature reuse (FR) as an innovative approach to expand the scope of model optimization for diffusion models. FR possesses distinct attributes compared to reduced NFE, enabling a synergistic effect when used together. In this chapter, we will explain the motivations behind the proposal of FR and discuss the expected advantages of this method.

\subsection{Temporal Redundancy}

Contemporary diffusion models often incorporate a series of blocks with a residual architecture. This design involves adding the layer's output to the input, as generally formulated by the following equation: 

\begin{equation}
\begin{aligned}
    \mathbf{y}^t_i & = \mathcal{F}_i(\mathbf{x}^t_i , t ) + \mathbf{x}^t_i.
    \label{eq:residual_t}
\end{aligned}
\end{equation}
Here, $i$ represents the index of layer, and  $\mathcal{F}(\cdot)$ denotes the layer function, $\mathbf{x},\mathbf{y}$ denote the input and output of the residual block, respectively, and $t \in [1, 2, ... , T]$ denotes the time step. Please note that   
Eq.\ref{eq:residual_t} incorporates temporal information as an input, allowing the model to be conditioned on time step. To generate high-quality images, the score estimation network $\epsilon(x_t,t)$ is repeatedly employed, taking the noisy image $x_t$ at time step $t$ as input and predicting the added noise.

The function $\mathcal{F}_i(\mathbf{x}^t_i , t )$ may take various forms depending on the architecture of diffusion model. For instance, in \cite{ldm}, $\mathcal{F}_i(\cdot)$ represents a convolutional layer or a self-attention layer with U-net structure, while in \cite{dit}, it resembles a ViT-like Block. The detailed specification of $\mathcal{F}_i(\cdot)$ in various diffusion model is provided in the Appendix. 

In this paper, our primarily focus on the temporal behavior of diffusion model, stemming from their repeated denoising operation. Specifically, we observe that the temporal changes of diffusion model remains relatively small across most time steps, regardless of the architecture and dataset. To aid the reader's comprehension, we offer a quantitative measure of \textit{temporal change} in $i$th layer of diffusion model, as described by the following equation:

\begin{equation}
    K_i(t,t') =   \expect{  \frac{\norm{\mathcal{F}_i(\mathbf{x}^t,t) - \mathcal{F}_i(\mathbf{x}^{t'},{t'})}{1}}{\Delta t} }{\mathbf{x}},
    \label{eq:temporal_change}
\end{equation}
where \( \Delta t= \|t-t'\| \). In Fig \ref{fig:temporal_similarity}(b), we showcase $K(t,t+1)$ across different layers of SDXL \cite{sdxl} and DiT \cite{dit}. As depicted, the temporal differences are minimal for the majority of time steps. We also provide visual evidence of this strong similarity in Fig \ref{fig:temporal_similarity}(a).  This similar appearance suggests that diffusion models may undergo redundant computations during the sampling process, indicating ample room for optimization.

\subsection{Feature Reuse}

\begin{figure}[t]
    \centering
        \includegraphics[width=\columnwidth]{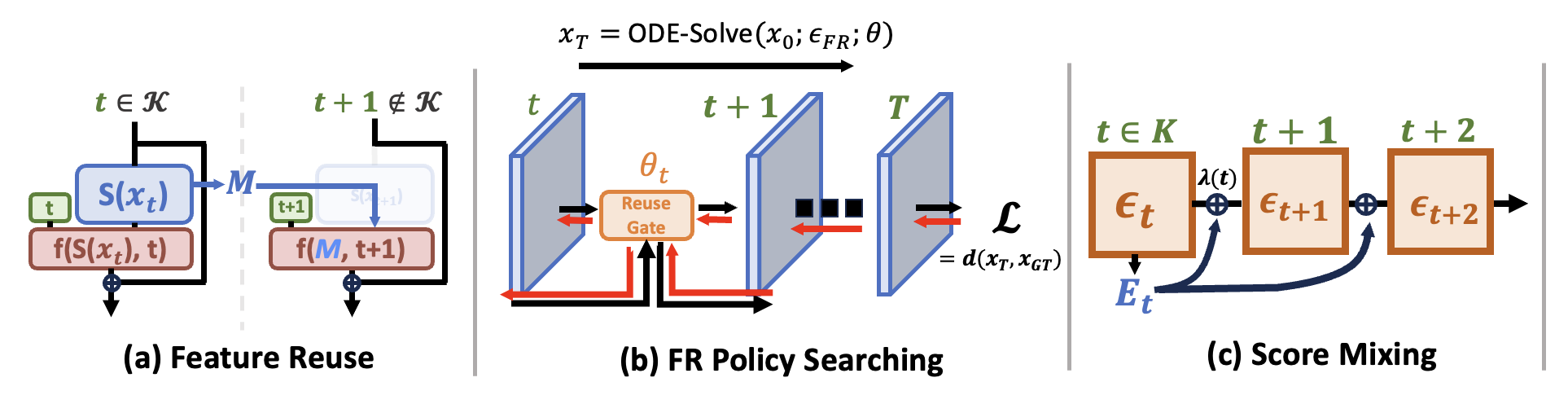}
    \caption{Overview of our methods. (a) Feature Reuse (Eq. \ref{eq:feature_reusing}) (b) Auto-FR for Optimal $\mathcal{K}$ Searching (Eq. \ref{eq:search_loss})  (c) Score Mixing (Eq. \ref{eq:mix})}
    \label{fig:overview}
\end{figure}

Expanding the earlier observation, we introduce a method to remove the unnecessary computations in the denoising process. Specifically, we store the results of intermediate features from previous timesteps and reuse them in the subsequent timesteps. While this idea is simple and intuitive, we analyze its effects in depth and propose a novel approach to maximize the benefits we can derive. 

First, we can decompose the computation of residual block into two parts:
\begin{equation}
     \mathcal{F}_i(\mathbf{x}^t_i , t ) = f_i(\mathcal{S}_i(\mathbf{x}^t_i) , t) ,
     \label{eq:res_break}
\end{equation}
where $\mathcal{S}(\cdot)$ is the operation performed before considering temporal information. 
Next, we define the  \textit{keyframe set} $\mathcal{K} \subseteq \{1, .. , N\} $, which represents the set of timesteps where the entire layers are updated and feature maps are saved for feature reusing. Here, $N$ denotes the number of sampling steps for generation (e.g., $N=50$ in DDPM). 

For \textit{keyframe} timestep $t \in \mathcal{K}$ , The result of $\mathcal{S}(\cdot)$ is stored in memory $\mathbf{M}^t_i$ for future reuse. The remaining operation in residual block is performed normally. This process can be expressed as follows:
\begin{equation}
\begin{aligned}
    \mathbf{M}^t_i & \leftarrow \mathcal{S}_i(\mathbf{x}^t_{i}),\\
    \mathbf{y}^t_i & = f_i(\mathbf{M}^t_i , t) + \mathbf{x}^t_i.
    \label{eq:feature_saving}
\end{aligned}
\end{equation}

For \textit{non-keyframe} timesteps $t' \notin \mathcal{K}$, the computation of the $\mathcal{S}(\cdot)$ is replaced by the saved memory from the nearest early timestep $t \in \mathcal{K}$, as follows:
\begin{equation}
\mathbf{y}^{t'}_i = f_i(\mathbf{M}^t_i , t') + \mathbf{x}^{t'}_i.
\label{eq:feature_reusing}
\end{equation}
Hence, by skipping the operations $\mathcal{S}(\cdot)$, a significant amount of computation can be saved, as illustrated in \cref{fig:overview} (a). Moreover, since the sequential denoising operation of the diffusion model typically progresses from $t=1$ to $N$, feature reuse can be implemented with a single memory $M_i$ for each layer. An important point to note is that \textbf{the temporal information is updated while $\mathcal{S}(\cdot)$ is reused}. This allows the time-conditioned information to propagate through the residual path, enabling the diffusion model to be conditioned on the correct time step. Because this feature reuse scheme is highly flexible, it can be applied to any diffusion model architecture that utilizes skip connection, including both U-Net and diffusion transformer architectures.

\subsection{Analysis: the effect of $\mathcal{K}$ selection }
The generation quality and acceleration effect of FR can vary significantly depending on the appropriate keyframe set. In this section, we analyze the effect of keyframe set selection using heuristic design. The simplest way to construct a keyframe set is to compose it as a collection of timesteps with uniform intervals. This $\mathcal{K}_{uniform}$ is defined as follows:
\begin{equation}
    \mathcal{K}^{M}_{uniform} = \{ t \in \mathbb{N}\ |\ t\ \text{mod}\ M =0 , t \leq N\} 
    \label{eq:unif_k}
\end{equation}
where \textbf{the FR interval} $M$ represents how long the saved data will be reused.

\subsubsection{Acceleration with Feature Reuse}

\begin{figure*}[t]
    \centering
    \begin{subfigure}[b]{0.32\columnwidth}
                \includegraphics[width=\textwidth]{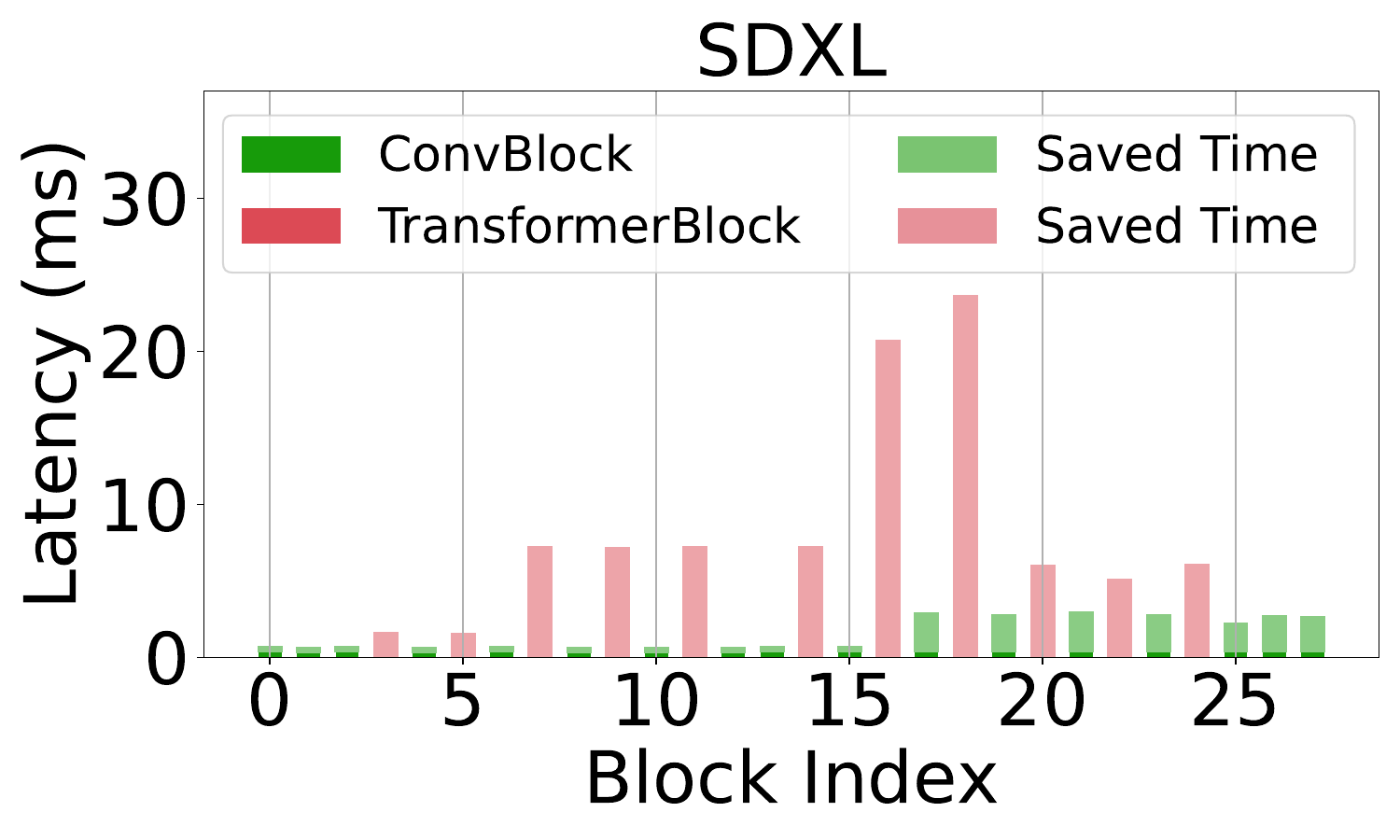}
                \caption{SDXL\cite{sdxl}}
    \end{subfigure}
    \begin{subfigure}[b]{0.32\columnwidth}
                \includegraphics[width=\textwidth]{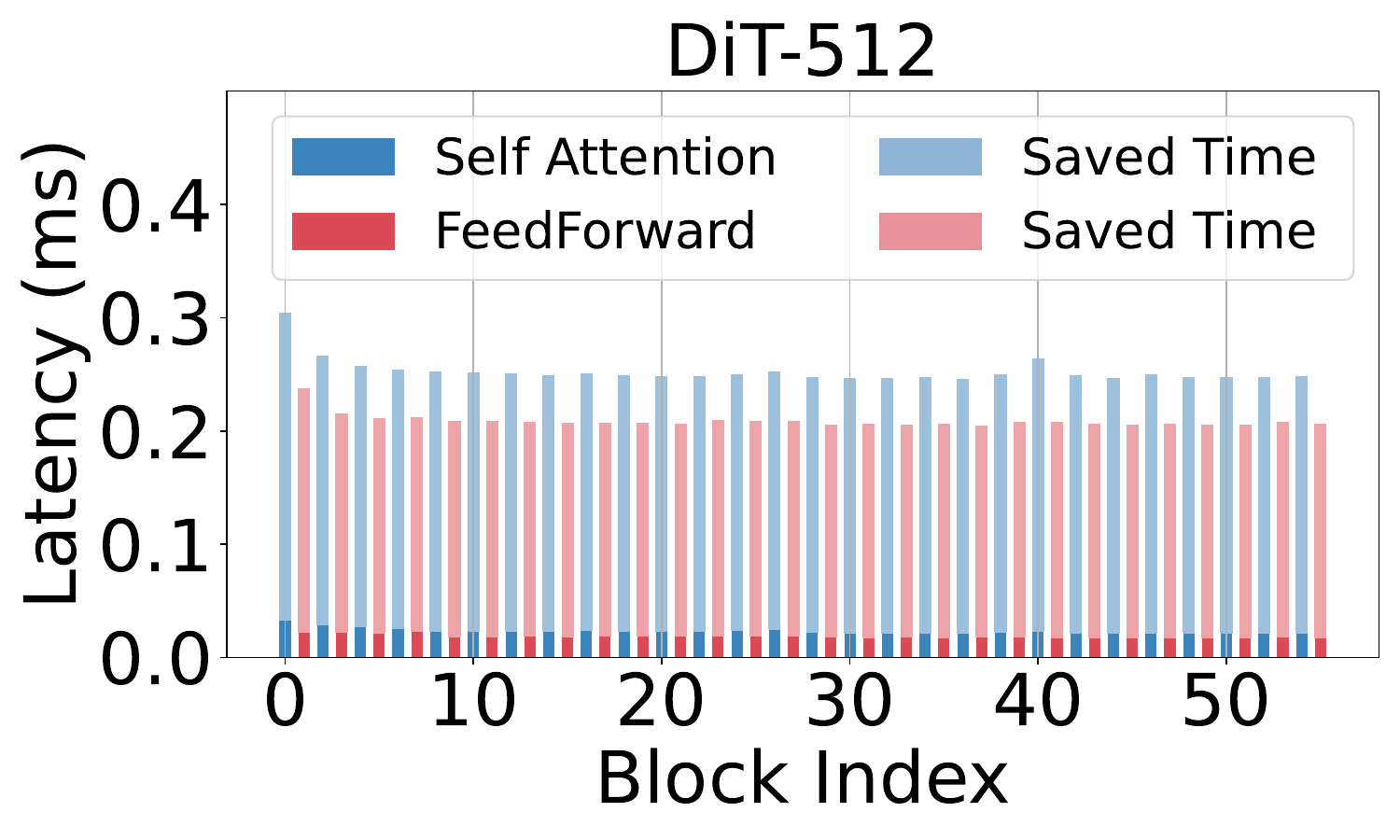}
                \caption{DiT-512\cite{dit}}
    \end{subfigure}
    \begin{subfigure}[b]{0.32\columnwidth}
                \includegraphics[width=\textwidth]{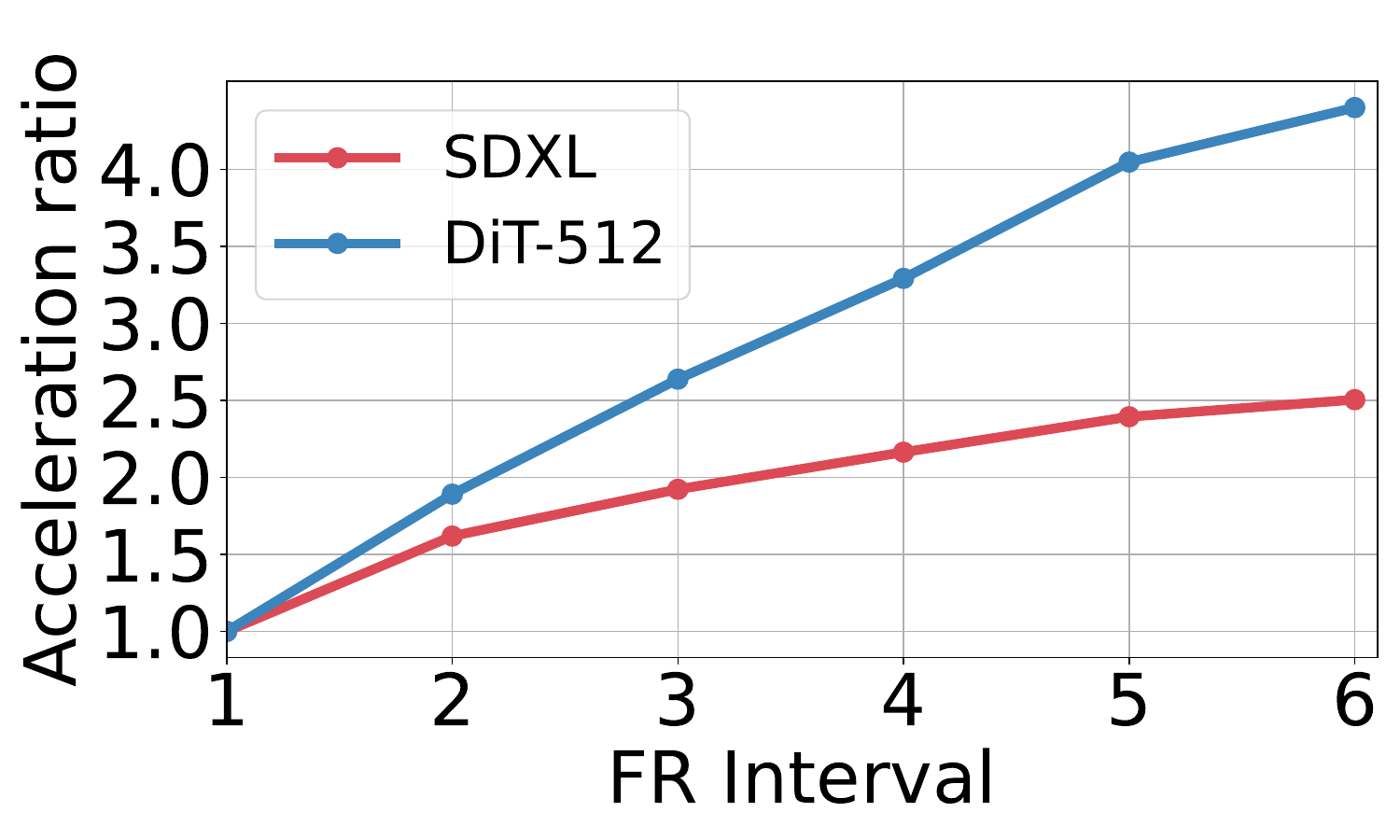}

                \caption{Acceleration Ratio}
    \end{subfigure}
    \caption{Detailed Analysis of Skippable Latency by Feature Reusing (FR) Technique for (a) SDXL (U-Net) and (b) DiT (Diffusion Transformer). (c) Speed-Up of FR Regarding FR Interval. }
    \label{fig:blockwise-latency}
\end{figure*}

Because we only reuse a portion of the Residual block, we need to validate whether this method can indeed offer practical advantages. Therefore, we conducted a comprehensive analysis to evaluate the latency with FR on a real device (Gefore RTX 3090).

In Fig. \ref{fig:blockwise-latency}, we depict the latency profiles of individual blocks within two different diffusion model architectures: U-Net(SDXL\cite{sdxl}) and DiT\cite{dit}. In the case of U-Net, the transformer block accounts for a larger portion of the execution cycle. This is primarily due to the spatial self-attention layer for high resolution features.  In the case of DiT, almost every block shows the same latency because it has consistent dimensions for every layer. For both architectures, we can save latency of approximately 92 percent and 68 percent, respectively, with FR.

In Fig. \ref{fig:blockwise-latency}(c), we measured the actual acceleration of FR against the FR interval. We use $N=50$ in this experiment. As shown in the figure, because there are parts that are not skipped, the speed up saturates as the FR interval increases. However, even a small FR interval, e.g., 2 to 3, offers notable improvement.

\subsubsection{Output Quality with Feature Reuse}
As explained in the previous section, FR could provide advantages in terms of both performance and quality. However, there are other acceleration methods such as the reduced NFE, so the adoption of FR should be justified by the distinctive advantages of FR over the reduced NFE. In this section, we will elucidate the unique benefits of FR we have discovered.

First, we explain the relation between the reduced NFE and FR. If we consider the very coarse-grained form of FR, which is reusing the entire output of the diffusion model $\epsilon(x,t)$, it is equivalent to the case of the reduced NFE. Indeed, FR is the fine-grained skipping in layer-wise granularity while the reduced NFE is the coarse-grained skipping in network-wise granularity. We provide a detailed interpretation of it for the DDIM case in the appendix.

\begin{figure}[t]
    \centering


    \begin{subfigure}[b]{0.345\columnwidth}
                \includegraphics[width=\textwidth]{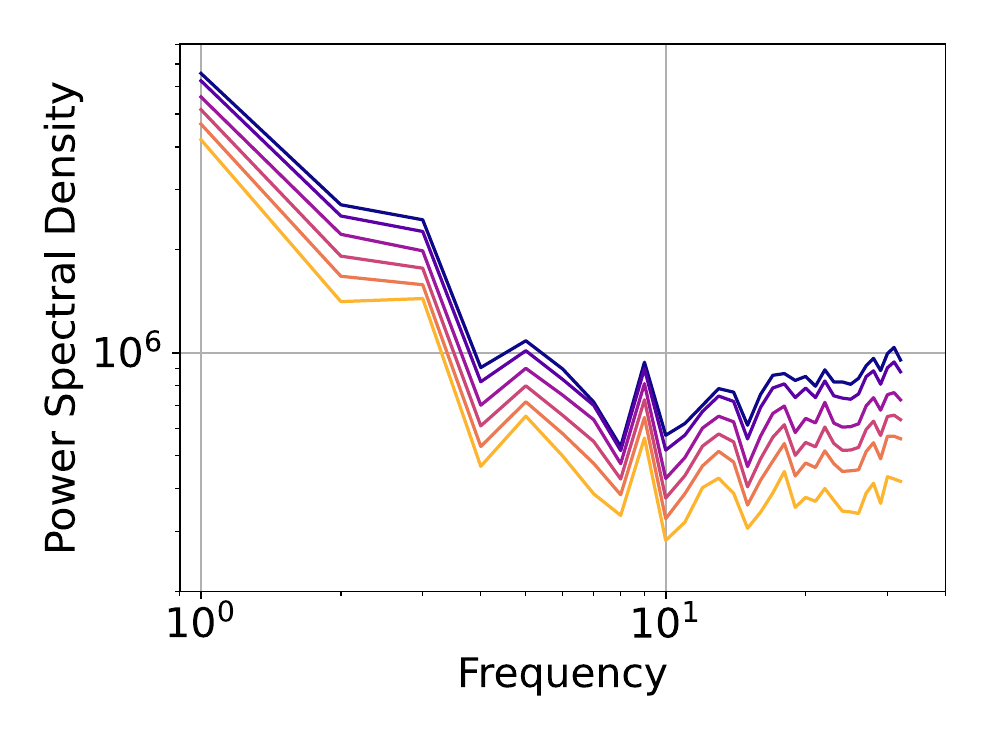}
                \caption{Reduced NFE}
    \end{subfigure}
    \begin{subfigure}[b]{0.27\columnwidth}
                \includegraphics[width=\textwidth]{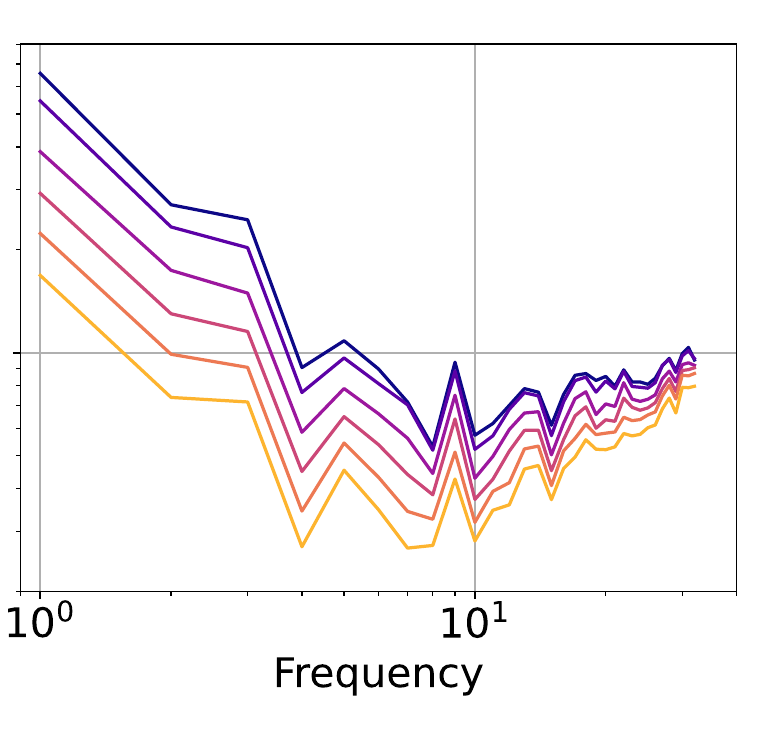}
                \caption{Feature Reuse}
    \end{subfigure}
     \begin{subfigure}[b]{0.35\columnwidth}
                \includegraphics[width=\textwidth]{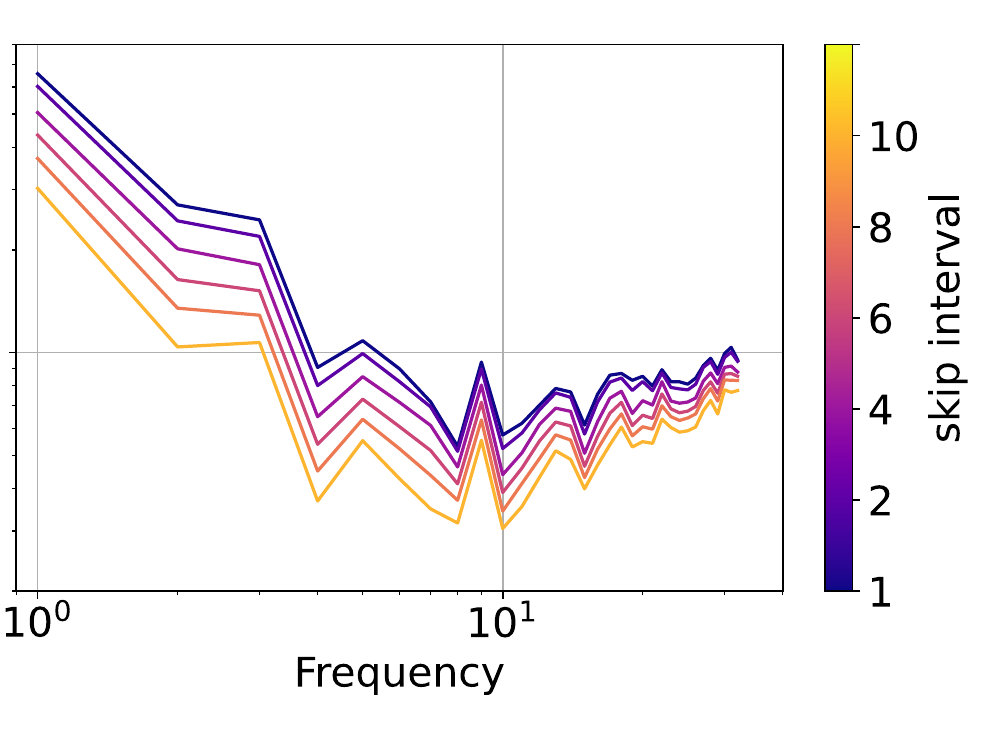}
                \caption{Score Mixing}
    \end{subfigure}

    \caption{Power Spectral Density (PSD) of Generated Image:
(a) Reduced NFE, (b) Naive Feature Reuse, and (c) The Proposed Idea, Score Mixing
    }
    \label{fig:PSD1}
\end{figure}

Intuitively, because the reduced NFE skips more computation than FR, this should generate more degraded output. However, we observe interesting patterns in terms of \textit{frequency response}. In Fig. \ref{fig:PSD1}, we depict the Power Spectral Density (PSD) analysis of generated images of the reduced NFE and FR. In the reduced NFE, the skip interval is increased from 1 to 10 while the FR interval ranges from 1 to 10 in FR. As shown in the figure, (a) the reduced NFE loses many high-frequency components while preserving the low-frequency area well. Meanwhile, (b) FR loses more low-frequency components while better preserving the high-frequency components. Please check the appendix for the visual comparison.

Our empirical findings suggest that FR is not consistently better than the reduced NFE; they possess distinct strengths and weaknesses. At this point, we need to pay attention to the recent findings on the generation characteristics. Recent studies \cite{ma2022accelerating, choi2022perception, diffusion-slim, diffprune} have shown that in the early denoising stages, the model mainly generates coarse-grained low-frequency components. In contrast, during the later stages, it predominantly generates fine-grained high-frequency components. Therefore, by combining these operational characteristics with our observations, \textbf{we devised a strategy that primarily employs the reduced NFE in the initial stages to generate coarse-grained structure and then switches to FR in the later stages to retain fine-grained details.}

\subsection{Score Mixing }

\subsubsection{Unified view of FR and the reduced NFE : Score mixing}
To integrate the two methods, we introduce an additional heuristic called ``score mixing''. Instead of just switching from the reduced NFE to FR, we propose to use mixture of the output of this two method. 
Specifically, at $t \in \mathcal{K}$, we also save the output of the model $\epsilon(x_t, t)$ to memory,
\begin{equation}
    E_t \leftarrow \epsilon(x_t,t)
\label{eq:score_saving}
\end{equation}
Then, we employ a linear interpolation of the score estimated by FR and output from previous keyframe, controlled by mixing schedule $\lambda(t)$. This modified score is inputted into the next iteration step of diffusion model.
\begin{gather}
    \epsilon(x_{t'},t') \leftarrow \lambda(t')*\epsilon(x_{t'},t') + (1-\lambda(t'))*E_t
    \label{eq:mix}\\ 
    \lambda(t) = \max(0, \min(1, (\tau*((t/N)-b) + 2) / 4 ) ).
\label{eq:hard_sig}
\end{gather}
While any increasing function can be used for $\lambda(t)$, we use the \textit{hard sigmoid} function. By using this $\lambda(t)$, we can skip the computation of the FR score when $\lambda(t) = 0$. In the case of conditional sampling (CFG), we simply mix the conditional score in the same way as the unconditional score, using the same $\lambda$.

In Eq. \ref{eq:hard_sig}, $\tau$ is the temperature that controls the switching speed of the schedule, and $b$ is the bias that controls the phase transition point of the schedule. We empirically determine the optimal values as $\tau=30$ and $b=0.5$ and use these values throughout the rest of the paper. In Fig.\ref{fig:PSD1}.(c), we depict the PSD analysis of the generated image using our score mixing. As shown in the figure, this exhibits the preservation of low and high frequencies compared to FR and the reduced NFE.

\subsection{Auto-tuning for FR interval: Auto-FR}
With score mixing, FR is used when $\lambda(t) > 0$. To maximize the benefit of FR, we propose an automated search called \textbf{Auto-FR} to find the optimal $\mathcal{K}$. In this approach, we apply a timestep-wise learnable parameter $\alpha_t =  sigmoid(\theta_t)$ with a hard gating mechanism update. It's important to note that the network parameters are frozen and training-free; only the gating parameters are updated. The forward path of the residual block is computed as follows:
\begin{gather}
    \alpha^*_t = \lfloor \alpha_t \rceil + \alpha_t - \texttt{stopgrad}(\alpha_t),
    \label{eq:ste_alpha}\\
    \mathbf{M}_i \leftarrow \alpha_t^* \cdot \mathcal{S}_i(\mathbf{x}^t_{i}) + (1-\alpha_t^*)\cdot \mathbf{M}_i. \quad
    \mathbf{y}^t_i = f_i(\mathbf{M}_i , t) + \mathbf{x}^t_i
    \label{eq:gating}
\end{gather}
Likewise, the forward path of score mixing is computed with gate parameter :
\begin{equation}
\begin{aligned}
    \mathbf{E} & \leftarrow \alpha_t^* \cdot \epsilon(x_t,t) + (1-\alpha_t^*)\cdot \mathbf{E}. \quad
    \epsilon(x_t,t) & = \lambda(t)*\epsilon(x_t,t) + (1-\lambda(t))*\mathbf{E}
\end{aligned}
\end{equation}

With this scheme, we can safely simulate and differentiate through the sampling process of FR if $t=0\rightarrow N$. We update $\theta_t$ using gradient descent with straight-through estimation to minimize the following loss function $\mathcal{L}(\cdot)$:
\begin{equation}
\begin{aligned}
    \mathcal{L}(\theta) = \expect{\norm{x^{GT}- \texttt{ODE-Solve}(x_0;N,\theta)}{2}}{x} + \lambda*\mathcal{L}_{cost}(\theta)
    \label{eq:search_loss}
\end{aligned}
\end{equation}
Here, $x^{GT}$ is the ground-truth sample generated from noise $x_0$, \texttt{ODE-Solve} is a differentiable ODE Solver (e.g., DDIM), $\mathcal{L}_{cost}$ is the latency cost of FR, and $\lambda$ is the balancing parameter that effectively controls the trade-off between latency and fidelity. The detailed training recipe is provided in the appendix. In short, the reuse policy is trained to maximize quality while minimizing the computation cost of the sampling process. 

Finally, the keyframe set is determined from the trained $\theta^*$:
\begin{equation}
    \mathcal{K}^{\lambda}_{search} = \{ t \in \mathbb{N}\ |\theta^*_t \geq 0, 0 \le t \leq N  \}.
    \label{eq:k_search}
\end{equation}
\textbf{Our final solution, FRDiff, is the mixture of score mixing and Auto-FR}, designed to leverage the benefits of the reduced NFE and FR with minimal human intervention.

\section{Experiments}
\label{sec:experiment}
\begin{figure*}[t!]
    \centering
        \includegraphics[width=\columnwidth]{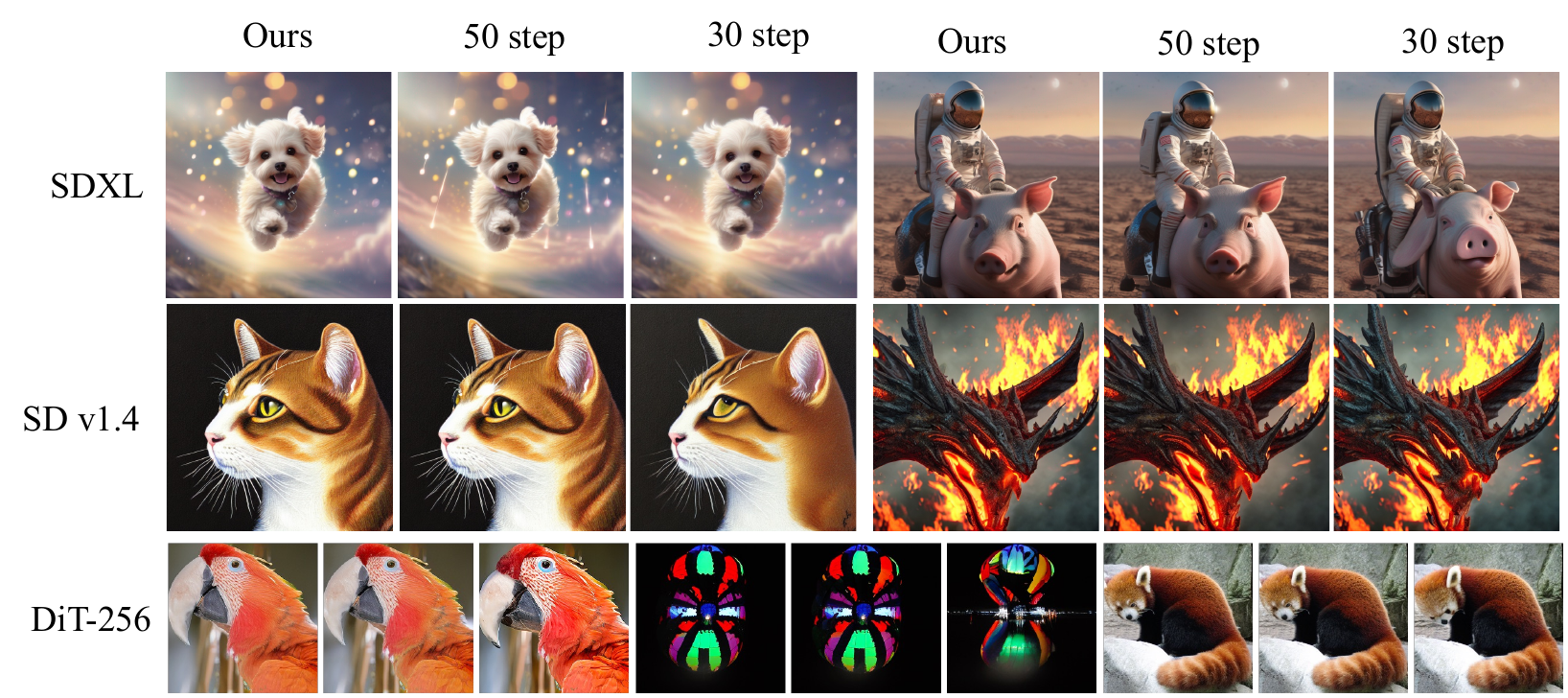}
    \caption{Qualitative Comparison of FRDiff. \textit{Best viewed zoomed-in}. (left) Ours  (middle) DDIM 50 step baseline (right) DDIM 30 step. Our FRDiff up to 1.76x faster than the baseline, DDIM 50.}
    \label{fig:qual_comparison}
\end{figure*}

\begin{table}[t!]
\centering
\small
\caption{Quantitative Results of FRDiff. }
\resizebox{0.83\columnwidth}{!}{
\label{tab:quant_comparison}
\begin{tabular}{c|c|c|ccc}
\toprule
Model & \; NFE \; & \; FRDiff \;& \; Latency(s)$\downarrow$ &\; Speed-up$\uparrow$ & \; FID$\downarrow$ \;\\
\midrule
\midrule

\multirow{3}{*}{CIFAR-10 \cite{ho2020denoising}} & 50 &            & 0.836 & 1.00x  & 4.03\\
                          & 30 &            & 0.495 & 1.68x & 5.01\\
                          & 50 & \Checkmark & \textbf{0.491} & \textbf{1.70x} & \textbf{4.64} \\
\midrule
\multirow{3}{*}{LDM-CelebA \cite{ldm}} & 50 &            & 1.317 & 1.00x  & 6.0\\
                            & 30 &            & 0.763 & 1.72x & 7.22\\
                            & 50 & \Checkmark & \textbf{0.748} & \textbf{1.76x}  & \textbf{6.33} \\
\midrule
\multirow{3}{*}{SD v1.4 \cite{ldm}} & 30 &  & 4.657 & 1.00x & 6.32 \\
                         & 20 &  & 3.027 & 1.53x & 8.41 \\
                         & 30 & \Checkmark & \textbf{2.947} & \textbf{1.58x} & \textbf{6.86} \\
\midrule
\multirow{3}{*}{SDXL \cite{sdxl}}    & 30 &  & 8.810 & 1.00x & 7.41 \\
                         & 20 &  & 6.033 & 1.46x & 9.46 \\
                         & 30 & \Checkmark & \textbf{5.491} & \textbf{1.60x} & \textbf{9.28} \\
\midrule
\multirow{3}{*}{DiT-256 \cite{dit}} & 30 & & 0.763 & 1.00x & 14.76 \\
                         & 20 & & 0.515 &  1.48x & 17.69 \\
                         & 30 & \Checkmark & \textbf{0.463} & \textbf{1.64x} & \textbf{16.71} \\
\bottomrule
\end{tabular}
}
\end{table}

\subsection{Experiments Setup}
To validate the effectiveness of our proposed idea, we assessed its efficacy across various existing diffusion models, including pixel-space (CIFAR-10) \cite{ho2020denoising}, latent diffusion model (LDM) \cite{ldm}, and Diffusion Transformer (DiT) \cite{dit}. Our intentional choice of diverse models aimed to demonstrate the versatility of the proposed idea. For example, the pixel-space model and LDM utilize a U-net structure, while DiT employs the diffusion transformer instead of U-net. Our method is designed to be applicable across all these model architectures. We obtained pretrained weights from the official repository for all models except for the pixel-space model (CIFAR-10).

For evaluation, we conducted both qualitative and quantitative experiments. In the qualitative assessment, we compared our generated images against the baseline images produced by DDIM with 50 steps \cite{ddim} in terms of fidelity and latency, as shown in Fig. \ref{fig:qual_comparison}. For quantitative evaluation, we measured the Fréchet Inception Distance (FID) \cite{fid} for Stable Diffusion(SD) \cite{ldm} and SDXL \cite{sdxl} using 5k samples in MS-COCO \cite{mscoco} and for DiT using 10k samples in ImageNet \cite{deng2009imagenet}. We also measure the speedup compared to the DDIM baseline, as summarized in Table \ref{tab:quant_comparison}. All experiments were conducted on a GPU server equipped with an NVIDIA GeForce RTX 3090, and latency measurements were performed using PyTorch \cite{PyTorch} with a batch size of 1 on a GeForce RTX 3090.

\subsubsection{Qualitative Analysis}
In Fig \ref{fig:qual_comparison}, we present a comparison of image generation results using various diffusion models: SDXL \cite{sdxl}, Stable Diffusion (SD) \cite{ldm}, and DiT-256 \cite{dit}. For comparison, we depict the baseline image (middle; DDIM with 50 steps), Our method (left; baseline + FRDiff), and DDIM with 30 steps, as fast as ours (right). As shown in the figure, Our method (left) can accelerate the baseline (middle) without quality degradation, while DDIM with reduced NFE (right) shows severe quality degradation, exhibiting notable artifacts. Our method can safely accelerate the diffusion sampling process up to 1.76x (average 1.62x) regardless of architecture and dataset.

\subsubsection{Quantitative Analysis}

In table \ref{tab:quant_comparison}, we also measured the FID score and relative speedup of baseline(DDIM 50), Ours(DDIM 50 + FRDiff), and the reduced NFE as fast as ours (DDIM 30) in various diffusion models. As shown in table, ours demonstrate superior performance than DDIM in terms of both latency and fidelity, regardless of dataset and architecture. For a more detailed analysis, please refer to the Pareto-line analysis, presented in Sec. \ref{sec:Pareto} and Fig. \ref{fig:new-pareto}.

\subsection{Comparison with Existing Methods}

In Table \ref{tab:comparison}, we compare our method with several recently developed fast sampling methods for diffusion models. Specifically, we categorize these methods into three types: distillation-based methods \cite{sdxlturbo, lcm}, advanced ODE solvers \cite{ddim, zheng2023dpmv3}, and other training-free acceleration methods \cite{deepcache}, which enable zero-shot acceleration. For distillation-based methods, although they achieve extremely low latency, their FID scores are relatively large, limiting their utility in cases where image quality is paramount. Additionally, the substantial training costs associated with these methods pose a significant barrier to widespread adoption. Next, we compare our method with recently proposed fast ODE solvers such as DDIM \cite{ddim} and DPM-Solver++ \cite{zheng2023dpmv3}. Our method demonstrates a smaller quality degradation compared to these recent ODE solvers, leveraging the unexplored potential of diffusion acceleration enabled by FR. Finally, we compare our method with DeepCache \cite{deepcache}, which also aims to accelerate the denoising process through feature caching. DeepCache saves intermediate activations in a depth-wise, coarse-grained manner. However, our FRDiff exhibits a better latency-tradeoff than DeepCache due to its fine-grained feature utilization and judicious design with score mixing and Auto-FR. Moreover, it's worth noting that while DeepCache's feature reusing scheme relies on UNet architecture, our method is applicable to any architecture that has a residual structure.

\begin{table}[t!]

\centering
\caption{Comparison of Existing Diffusion Acceleration Methods. }\label{tab:comparison}
\resizebox{0.7\columnwidth}{!}{
\begin{tabular}{c|c|cc|c}
\toprule
Method & \; NFE \; & \; Latency$\downarrow$  & \; Retrain & \; FID$\downarrow$ \; \\
\midrule
\midrule
SDXL-Turbo \cite{sdxlturbo} & 4 & 0.731  & \Checkmark & 22.58 \\
LCM \cite{lcm} & 1 & 0.171  & \Checkmark& 42.53 \\
\midrule
\midrule
\multirow{2}{*}{DDIM \cite{ddim}} & 30 & 4.654 &  \XSolidBrush & 6.31 \\
                                            & 20 & 3.078 &  \XSolidBrush & 8.45 \\
\midrule

\multirow{1}{*}{DPM-Solver++ \cite{zheng2023dpmv3}}
                                                    & 20 & 3.075 & \XSolidBrush & 6.63 \\
\midrule
\midrule
DeepCache \cite{deepcache} & 50 & 5.027 & \XSolidBrush & 6.34 \\
\midrule
\midrule
Ours (M=2) & 50 & 4.183 & \XSolidBrush & 6.40 \\
Ours (M=3) & 50 & 3.117 & \XSolidBrush & 7.60 \\
Ours (M=2) & 40 & 3.333 & \XSolidBrush & 7.24  \\
Ours (M=2) & 30 & 2.492 & \XSolidBrush & 7.83 \\
\midrule
Ours w/ AutoFR & 35 & 2.914 &  \textbf{0.1h} & \textbf{6.20} \\ 
\bottomrule
\end{tabular}
}
\end{table}

\subsection{Ablation Study}

\begin{figure}[t]
\centering
  \begin{minipage}{0.8\columnwidth}
    \centering
            \begin{subfigure}[b]{0.45\columnwidth}
            \centering
            \includegraphics[width=\textwidth]{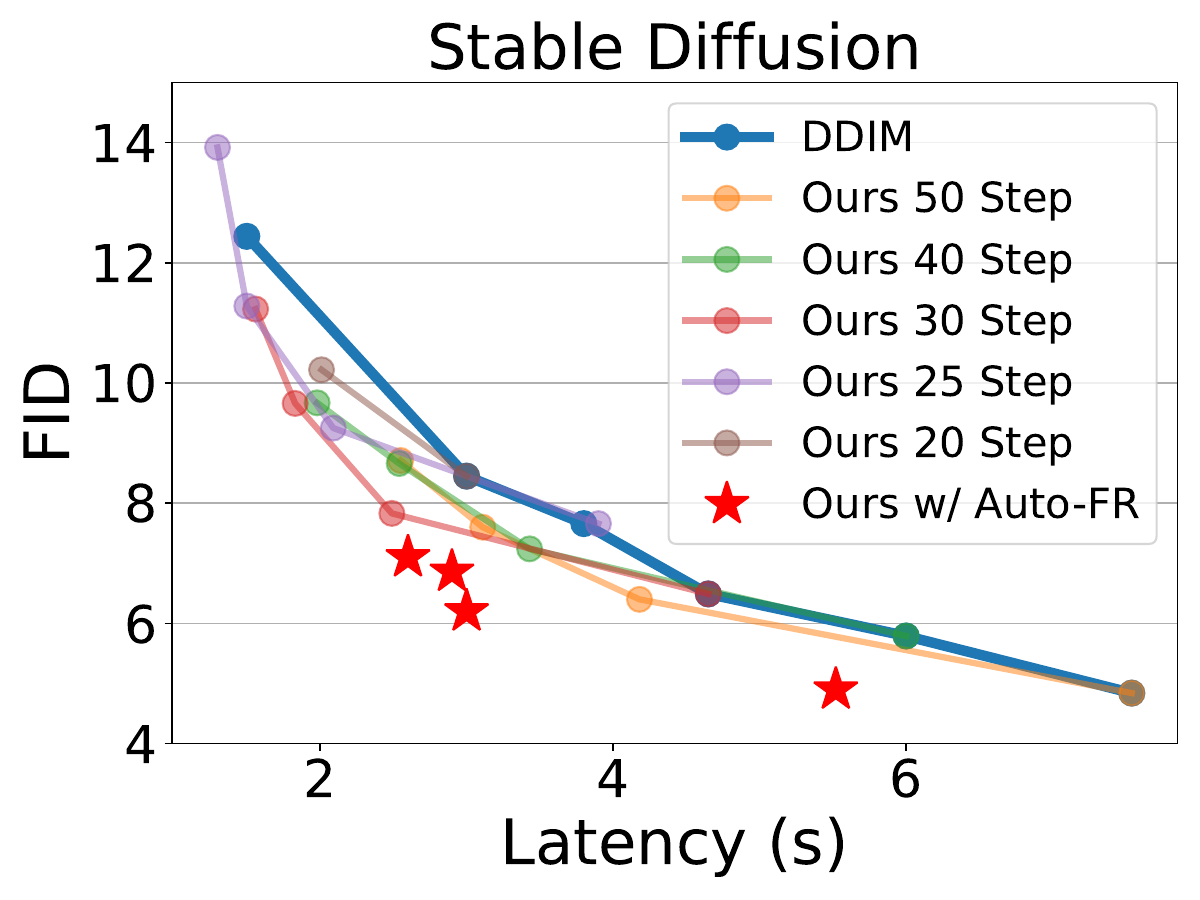}
            \caption{Stable-Diffusion}
        \end{subfigure}
        ~ 
        \begin{subfigure}[b]{0.45\columnwidth}
            \centering
            \includegraphics[width=\textwidth]{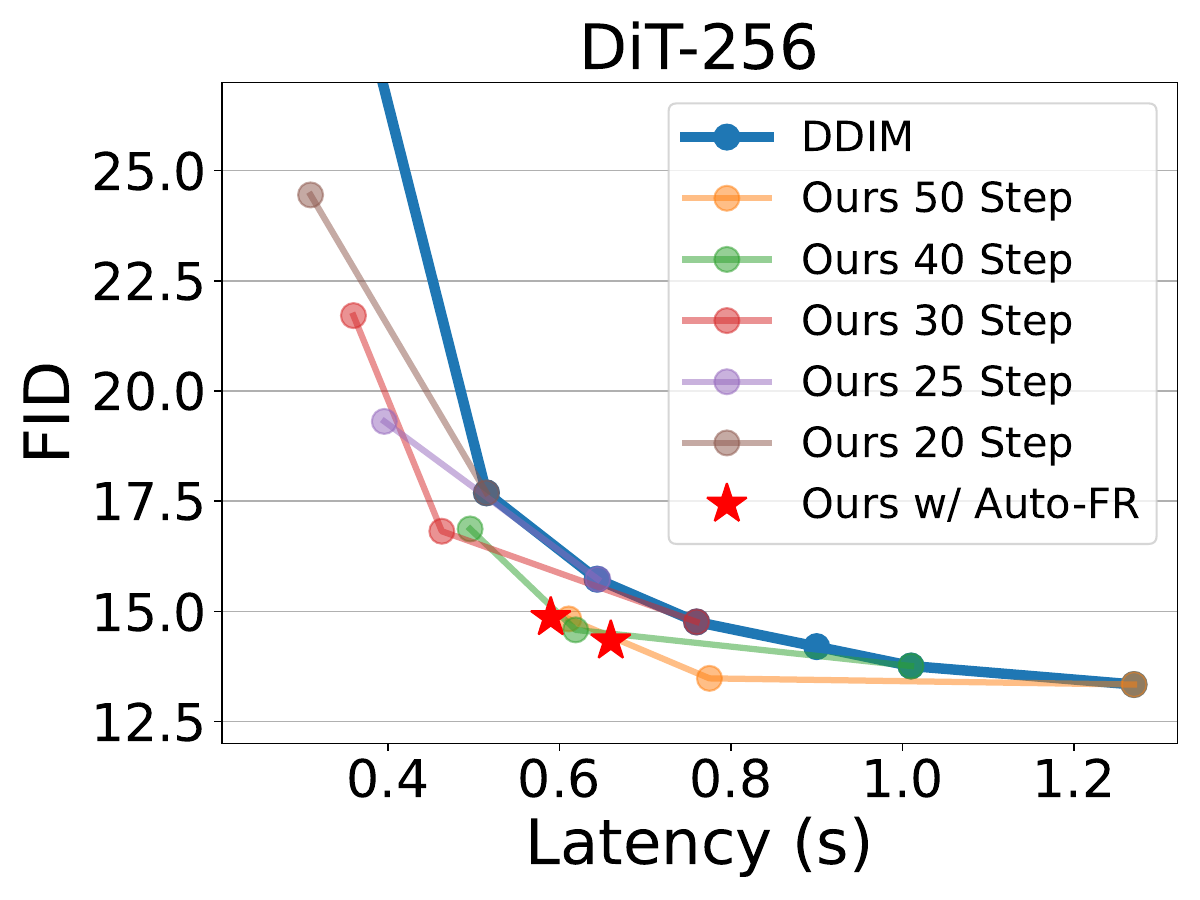}
            \caption{DiT-256}
        \end{subfigure}
        \caption{Pareto-line Comparison of the reduced NFE vs FRDiff}
        \label{fig:new-pareto}
        
  \end{minipage}
\end{figure}
\subsubsection{reducing NFE vs FR}\label{sec:Pareto}

In this section, we visualize the effect of changing the NFE and FR Interval (M) on SDXL \cite{sdxl} in Fig. \ref{fig:grid-result}. As depicted, reducing the NFE leads to a rapid degradation in performance, whereas reducing the Keyframe ratio maintains performance relatively well. Although FR incurs slightly more computation than reduced NFE, it offers new opportunities in the quality-latency trade-off.

To explore the trade-off relationship in detail, we draw Pareto lines in terms of latency and FID. In this experiment, the blue solid line represents the DDIM results with only reduced NFE. Meanwhile, our method adjusts both NFE and FR interval simultaneously; each line represents the corresponding reduced NFE, while the data point is added by increasing the FR interval gradually. As shown in the figure, adjusting both the keyframe ratio and NFE (Ours) clearly shows better Pareto fronts than DDIM. Furthermore, in Fig. \ref{fig:new-pareto}, we depict the points obtained by Auto-FR as $\star$ with different cost objectives. These points exhibit superior latency-FID Pareto fronts, illustrating the benefits of autoML-based optimization. We provide a detailed training recipe and trained keyframe sets of these points in the appendix. Our Auto-FR can find an optimal FR policy with minimal training costs, enabling us to exploit the benefits of FR with minimal human intervention.

\begin{figure*}[t]
    \centering
    \begin{minipage}{0.76\textwidth}
        \centering

        \begin{subfigure}[b]{0.45\textwidth}
          \includegraphics[width=\textwidth]{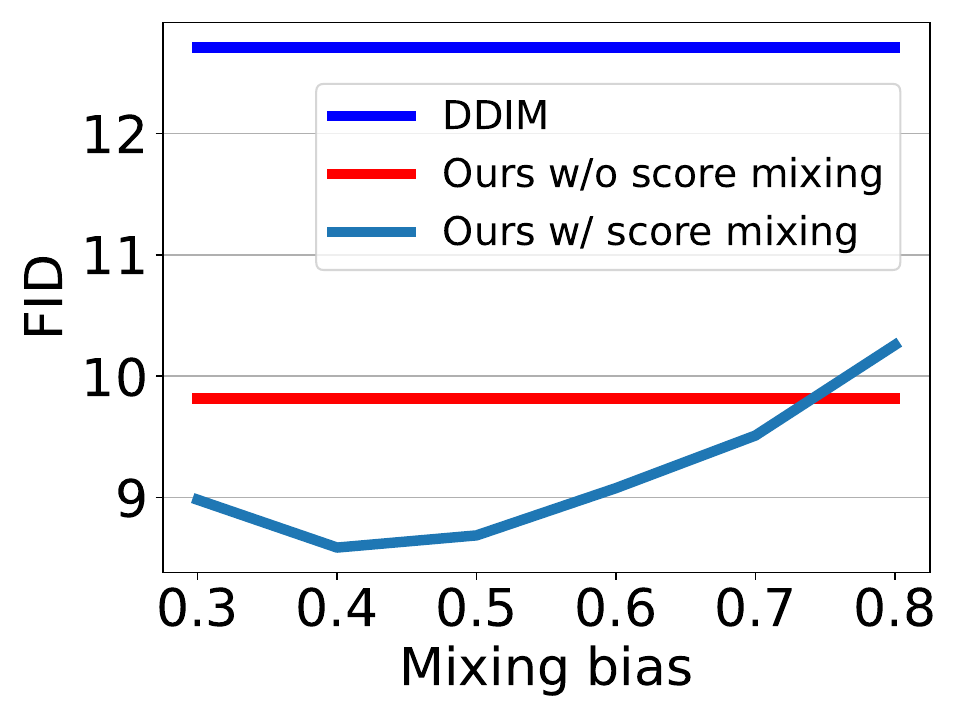}
          \caption{Skip interval 4}
          \label{fig:2a}
        \end{subfigure}
        \quad 
        \begin{subfigure}[b]{0.45\textwidth}
          \includegraphics[width=\textwidth]{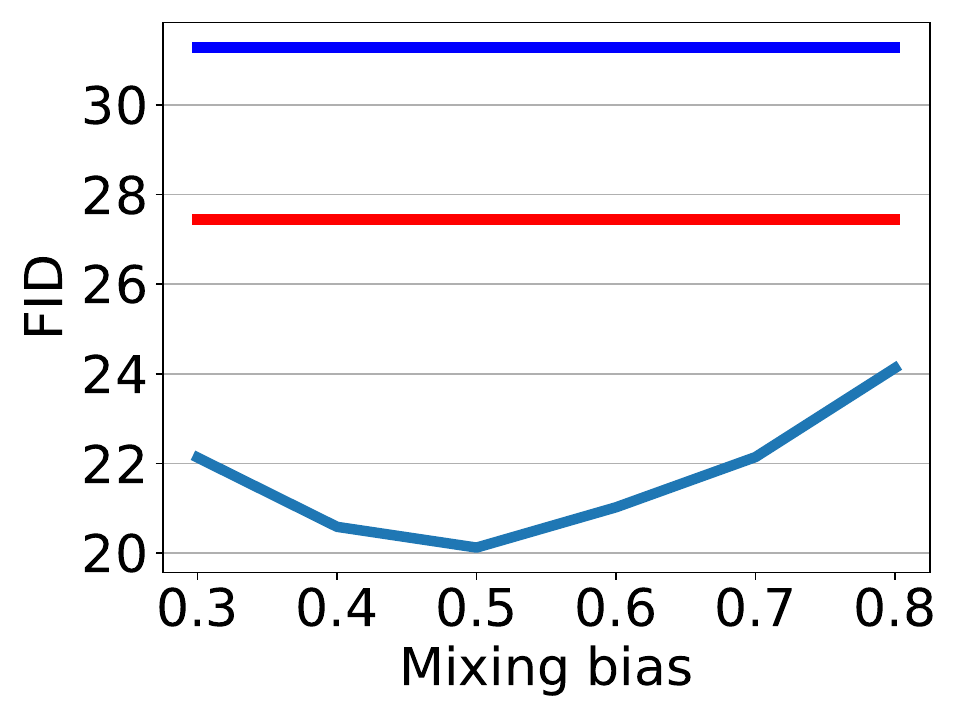}
          \caption{Skip interval 8}
          \label{fig:2b}
        \end{subfigure}
        \caption{Impact of Bias Value on FID in Score Mixing.}
        \label{fig:ablation-bias}
    \end{minipage}
    \begin{minipage}{0.22\textwidth}
            \centering
            \captionof{table}{Memory Overhead}
            \label{tab:feature-maps-size}
            \resizebox{\columnwidth}{!}{
                \begin{tabular}{lcc}
                    \toprule
                    \textbf{Model} & Size (MB) \\
                    \midrule
                    CIFAR-10        & 5.4 \\
                    LDM-4           & 44.4 \\
                    SD v1.4 & 180.0 \\
                    SDXL & 530.0 \\
                    DiT-512 & 252.0 \\
                    \bottomrule
                \end{tabular}
            }       
            
    \end{minipage}
\end{figure*}

\subsubsection{Score Mixing}
In Eq. \ref{eq:hard_sig}, the bias $b$ determines the transition point from the reduced NFE for low-frequency components to FR for high-frequency ones. To understand the impact of this bias on generation performance, we swept the bias from 0.3 to 0.8 and measured the FID score, as shown in Fig. \ref{fig:ablation-bias}. The figure illustrates that the optimal FID is achieved when the bias is around 0.4 to 0.5, indicating that the mixture of the reduced NFE and FR is definitely helpful in improving accuracy, with the advantageous interval being approximately half and half. Based on this observation, we empirically set $b=0.5$ throughout the rest of the paper.  

\begin{figure}[t!]
    \centering
        \includegraphics[width=0.99\columnwidth]{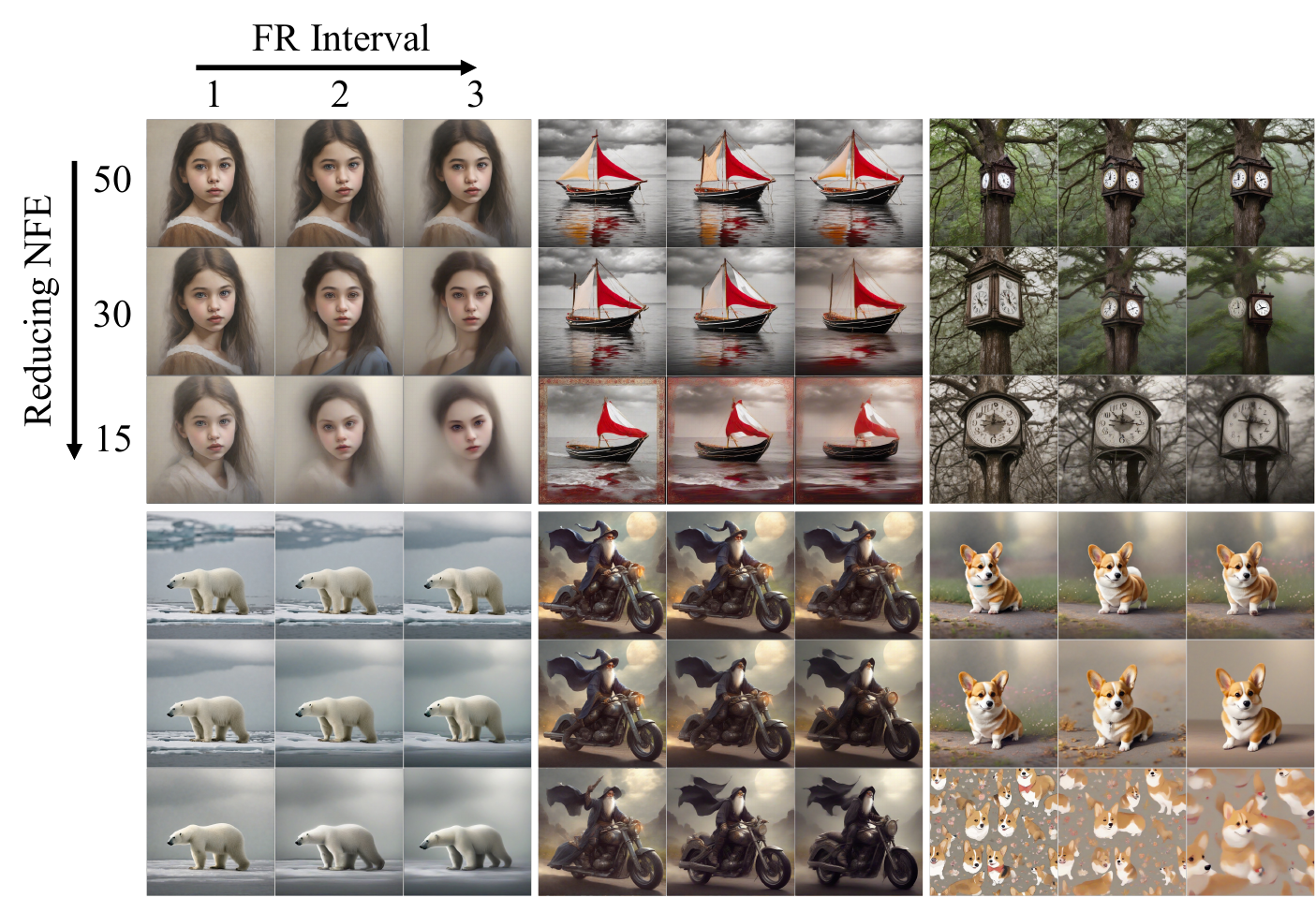}
    \caption{Visualization of Reduced NFE vs FR. \textit{Best viewed zoomed-in}. }
    \label{fig:grid-result}
\end{figure}

\subsubsection{Memory Overhead} 
FR needs to save the intermediate features for future timesteps, necessitating additional memory space. Table \ref{tab:feature-maps-size} presents the total amount of memory required for feature reusing. As shown in the table, our method can be applicable within an affordable memory overhead. For instance, in DiT-512, the model takes 4.2 GB of running memory; only adding 5.8\% of additional space, we enjoy 1.64x faster generation speed. 

\subsubsection{Other Tasks}
To assess the versatility of our approach, we apply it to several other tasks including super-resolution, image inpainting, and text-to-video generation. FRDiff proved to be applicable across all tasks, safely accelerating existing denoising processes. Additional results are provided in the Appendix.

\section{Discussion}
In this section, we will address the limitations of our approach. Although our method offers the advantage of plug-and-play integration without reliance on a particular ODE solver, its applicability becomes non-trivial when the time step for score function evaluation is not continuously provided. For example, in DPM-Solver++~\cite{lu2022dpmv2}, where 2 or 3 non-consecutive score function evaluations are needed to compute a single score, the efficacy of FR may diminish due to non-consecutive feature maps. Further investigation is warranted for such methods.

\section{Conclusion}
In this paper, we introduce \textit{FRDiff}, a novel Feature Reusing (FR)-based zero-shot acceleration technique for diffusion models. By leveraging the temporal similarity inherent in the iterative generation process, FRDiff can achieve remarkable acceleration of up to 1.76x without sacrificing output quality. Moreover, through a comprehensive examination, we present two additional techniques: score mixing, which harnesses the advantages of both reduced NFE and FR, and Auto-FR, which determines the optimal configuration through automated tuning. We validate our approach across diverse task datasets, demonstrating superior generation quality compared to existing acceleration methods within the same latency constraints.\\

\noindent \textbf{Acknowledgments}
This work was supported by IITP grant funded by the Korea
government (RS-2023-00228970, RS-2021-II210310, RS-2021-II210105, and RS-2019-II191906) and Samsung Research Global AI Center.
%
%
\bibliographystyle{splncs04}
\bibliography{main}

\begin{thebibliography}{10}
\providecommand{\url}[1]{\texttt{#1}}
\providecommand{\urlprefix}{URL }
\providecommand{\doi}[1]{https://doi.org/#1}

\bibitem{universal-guidance}
Bansal, A., Chu, H.M., Schwarzschild, A., Sengupta, S., Goldblum, M., Geiping, J., Goldstein, T.: Universal guidance for diffusion models. In: CVPR. pp. 843--852 (2023)

\bibitem{tome}
Bolya, D., Hoffman, J.: Token merging for fast stable diffusion. In: Proceedings of the IEEE/CVF Conference on Computer Vision and Pattern Recognition. pp. 4598--4602 (2023)

\bibitem{ip2p}
Brooks, T., Holynski, A., Efros, A.A.: Instructpix2pix: Learning to follow image editing instructions. In: CVPR. pp. 18392--18402 (2023)

\bibitem{choi2022perception}
Choi, J., Lee, J., Shin, C., Kim, S., Kim, H., Yoon, S.: Perception prioritized training of diffusion models. In: Proceedings of the IEEE/CVF Conference on Computer Vision and Pattern Recognition. pp. 11472--11481 (2022)

\bibitem{deng2009imagenet}
Deng, J., Dong, W., Socher, R., Li, L.J., Li, K., Fei-Fei, L.: Imagenet: A large-scale hierarchical image database. In: 2009 IEEE conference on computer vision and pattern recognition. pp. 248--255. Ieee (2009)

\bibitem{diffprune}
Fang, G., Ma, X., Wang, X.: Structural pruning for diffusion models. arXiv preprint arXiv:2305.10924  (2023)

\bibitem{mcm-diffusion}
Ham, C., Hays, J., Lu, J., Singh, K.K., Zhang, Z., Hinz, T.: Modulating pretrained diffusion models for multimodal image synthesis. SIGGRAPH Conference Proceedings  (2023)

\bibitem{dds}
Hertz, A., Aberman, K., Cohen-Or, D.: Delta denoising score. In: ICCV. pp. 2328--2337 (2023)

\bibitem{fid}
Heusel, M., Ramsauer, H., Unterthiner, T., Nessler, B., Hochreiter, S.: Gans trained by a two time-scale update rule converge to a local nash equilibrium. Advances in neural information processing systems  \textbf{30} (2017)

\bibitem{ho2020denoising}
Ho, J., Jain, A., Abbeel, P.: Denoising diffusion probabilistic models. Advances in neural information processing systems  \textbf{33},  6840--6851 (2020)

\bibitem{ho2022classifier}
Ho, J., Salimans, T.: Classifier-free diffusion guidance. arXiv preprint arXiv:2207.12598  (2022)

\bibitem{celeba-hq}
Karras, T., Aila, T., Laine, S., Lehtinen, J.: Progressive growing of gans for improved quality, stability, and variation. arXiv preprint arXiv:1710.10196  (2017)

\bibitem{edm}
Karras, T., Aittala, M., Aila, T., Laine, S.: Elucidating the design space of diffusion-based generative models. Advances in Neural Information Processing Systems  \textbf{35},  26565--26577 (2022)

\bibitem{bksdm}
Kim, B.K., Song, H.K., Castells, T., Choi, S.: On architectural compression of text-to-image diffusion models. arXiv preprint arXiv:2305.15798  (2023)

\bibitem{kingma2013auto}
Kingma, D.P., Welling, M.: Auto-encoding variational bayes. arXiv preprint arXiv:1312.6114  (2013)

\bibitem{fasterdiffusion}
Li, S., Hu, T., Khan, F.S., Li, L., Yang, S., Wang, Y., Cheng, M.M., Yang, J.: Faster diffusion: Rethinking the role of unet encoder in diffusion models. arXiv preprint arXiv:2312.09608  (2023)

\bibitem{qdiff}
Li, X., Liu, Y., Lian, L., Yang, H., Dong, Z., Kang, D., Zhang, S., Keutzer, K.: Q-diffusion: Quantizing diffusion models. In: Proceedings of the IEEE/CVF International Conference on Computer Vision. pp. 17535--17545 (2023)

\bibitem{mscoco}
Lin, T.Y., Maire, M., Belongie, S., Hays, J., Perona, P., Ramanan, D., Doll{\'a}r, P., Zitnick, C.L.: Microsoft coco: Common objects in context. In: Computer Vision--ECCV 2014: 13th European Conference, Zurich, Switzerland, September 6-12, 2014, Proceedings, Part V 13. pp. 740--755. Springer (2014)

\bibitem{diffbir}
Lin, X., He, J., Chen, Z., Lyu, Z., Fei, B., Dai, B., Ouyang, W., Qiao, Y., Dong, C.: Diffbir: Towards blind image restoration with generative diffusion prior. arXiv preprint arXiv:2308.15070  (2023)

\bibitem{pndm}
Liu, L., Ren, Y., Lin, Z., Zhao, Z.: Pseudo numerical methods for diffusion models on manifolds. arXiv preprint arXiv:2202.09778  (2022)

\bibitem{liu2022flow}
Liu, X., Gong, C., Liu, Q.: Flow straight and fast: Learning to generate and transfer data with rectified flow. arXiv preprint arXiv:2209.03003  (2022)

\bibitem{lu2022dpm}
Lu, C., Zhou, Y., Bao, F., Chen, J., Li, C., Zhu, J.: Dpm-solver: A fast ode solver for diffusion probabilistic model sampling in around 10 steps. Advances in Neural Information Processing Systems  \textbf{35},  5775--5787 (2022)

\bibitem{lu2022dpmv2}
Lu, C., Zhou, Y., Bao, F., Chen, J., Li, C., Zhu, J.: Dpm-solver++: Fast solver for guided sampling of diffusion probabilistic models. arXiv preprint arXiv:2211.01095  (2022)

\bibitem{latentconsistency}
Luo, S., Tan, Y., Huang, L., Li, J., Zhao, H.: Latent consistency models: Synthesizing high-resolution images with few-step inference. arXiv preprint arXiv:2310.04378  (2023)

\bibitem{lcm}
Luo, S., Tan, Y., Huang, L., Li, J., Zhao, H.: Latent consistency models: Synthesizing high-resolution images with few-step inference. arXiv preprint arXiv:2310.04378  (2023)

\bibitem{ma2022accelerating}
Ma, H., Zhang, L., Zhu, X., Feng, J.: Accelerating score-based generative models with preconditioned diffusion sampling. In: European Conference on Computer Vision. pp. 1--16. Springer (2022)

\bibitem{deepcache}
Ma, X., Fang, G., Wang, X.: Deepcache: Accelerating diffusion models for free. In: Proceedings of the IEEE/CVF Conference on Computer Vision and Pattern Recognition. pp. 15762--15772 (2024)

\bibitem{meng2023distillation}
Meng, C., Rombach, R., Gao, R., Kingma, D., Ermon, S., Ho, J., Salimans, T.: On distillation of guided diffusion models. In: Proceedings of the IEEE/CVF Conference on Computer Vision and Pattern Recognition. pp. 14297--14306 (2023)

\bibitem{t2i-adapter}
Mou, C., Wang, X., Xie, L., Zhang, J., Qi, Z., Shan, Y., Qie, X.: T2i-adapter: Learning adapters to dig out more controllable ability for text-to-image diffusion models. arXiv preprint arXiv:2302.08453  (2023)

\bibitem{PyTorch}
Paszke, A., Gross, S., Massa, F., Lerer, A., Bradbury, J., Chanan, G., Killeen, T., Lin, Z., Gimelshein, N., Antiga, L., et~al.: Pytorch: An imperative style, high-performance deep learning library. Advances in neural information processing systems  \textbf{32} (2019)

\bibitem{dit}
Peebles, W., Xie, S.: Scalable diffusion models with transformers. In: Proceedings of the IEEE/CVF International Conference on Computer Vision. pp. 4195--4205 (2023)

\bibitem{sdxl}
Podell, D., English, Z., Lacey, K., Blattmann, A., Dockhorn, T., M{\"u}ller, J., Penna, J., Rombach, R.: Sdxl: Improving latent diffusion models for high-resolution image synthesis. arXiv preprint arXiv:2307.01952  (2023)

\bibitem{dalle}
Ramesh, A., Pavlov, M., Goh, G., Gray, S., Voss, C., Radford, A., Chen, M., Sutskever, I.: Zero-shot text-to-image generation. In: International Conference on Machine Learning. pp. 8821--8831. PMLR (2021)

\bibitem{preditor}
Ravi, H., Kelkar, S., Harikumar, M., Kale, A.: Preditor: Text guided image editing with diffusion prior. arXiv preprint arXiv:2302.07979  (2023)

\bibitem{ldm}
Rombach, R., Blattmann, A., Lorenz, D., Esser, P., Ommer, B.: High-resolution image synthesis with latent diffusion models. In: CVPR. pp. 10684--10695 (2022)

\bibitem{imagen}
Saharia, C., Chan, W., Saxena, S., Li, L., Whang, J., Denton, E.L., Ghasemipour, K., Gontijo~Lopes, R., Karagol~Ayan, B., Salimans, T., et~al.: Photorealistic text-to-image diffusion models with deep language understanding. NeurIPS  \textbf{35},  36479--36494 (2022)

\bibitem{salimans2022progressive}
Salimans, T., Ho, J.: Progressive distillation for fast sampling of diffusion models. arXiv preprint arXiv:2202.00512  (2022)

\bibitem{sdxlturbo}
Sauer, A., Lorenz, D., Blattmann, A., Rombach, R.: Adversarial diffusion distillation. arXiv preprint arXiv:2311.17042  (2023)

\bibitem{tdq}
So, J., Lee, J., Ahn, D., Kim, H., Park, E.: Temporal dynamic quantization for diffusion models. arXiv preprint arXiv:2306.02316  (2023)

\bibitem{sohl2015deep}
Sohl-Dickstein, J., Weiss, E., Maheswaranathan, N., Ganguli, S.: Deep unsupervised learning using nonequilibrium thermodynamics. In: International conference on machine learning. pp. 2256--2265. PMLR (2015)

\bibitem{ddim}
Song, J., Meng, C., Ermon, S.: Denoising diffusion implicit models. arXiv preprint arXiv:2010.02502  (2020)

\bibitem{song2023consistency}
Song, Y., Dhariwal, P., Chen, M., Sutskever, I.: Consistency models. arXiv preprint arXiv:2303.01469  (2023)

\bibitem{song2020score}
Song, Y., Sohl-Dickstein, J., Kingma, D.P., Kumar, A., Ermon, S., Poole, B.: Score-based generative modeling through stochastic differential equations. arXiv preprint arXiv:2011.13456  (2020)

\bibitem{pnp}
Tumanyan, N., Geyer, M., Bagon, S., Dekel, T.: Plug-and-play diffusion features for text-driven image-to-image translation. In: CVPR. pp. 1921--1930 (2023)

\bibitem{stablesr}
Wang, J., Yue, Z., Zhou, S., Chan, K.C., Loy, C.C.: Exploiting diffusion prior for real-world image super-resolution. arXiv preprint arXiv:2305.07015  (2023)

\bibitem{mdp}
Wang, Q., Zhang, B., Birsak, M., Wonka, P.: Mdp: A generalized framework for text-guided image editing by manipulating the diffusion path. arXiv preprint arXiv:2303.16765  (2023)

\bibitem{boxdiff}
Xie, J., Li, Y., Huang, Y., Liu, H., Zhang, W., Zheng, Y., Shou, M.Z.: Boxdiff: Text-to-image synthesis with training-free box-constrained diffusion. In: ICCV. pp. 7452--7461 (2023)

\bibitem{xu2023restart}
Xu, Y., Deng, M., Cheng, X., Tian, Y., Liu, Z., Jaakkola, T.: Restart sampling for improving generative processes. arXiv preprint arXiv:2306.14878  (2023)

\bibitem{lawdiffusion}
Yang, B., Luo, Y., Chen, Z., Wang, G., Liang, X., Lin, L.: Law-diffusion: Complex scene generation by diffusion with layouts. In: Proceedings of the IEEE/CVF International Conference on Computer Vision. pp. 22669--22679 (2023)

\bibitem{diffusion-slim}
Yang, X., Zhou, D., Feng, J., Wang, X.: Diffusion probabilistic model made slim. In: Proceedings of the IEEE/CVF Conference on Computer Vision and Pattern Recognition. pp. 22552--22562 (2023)

\bibitem{freedom}
Yu, J., Wang, Y., Zhao, C., Ghanem, B., Zhang, J.: Freedom: Training-free energy-guided conditional diffusion model. ICCV  (2023)

\bibitem{controlnet}
Zhang, L., Rao, A., Agrawala, M.: Adding conditional control to text-to-image diffusion models. In: CVPR. pp. 3836--3847 (2023)

\bibitem{gddim}
Zhang, Q., Tao, M., Chen, Y.: gddim: Generalized denoising diffusion implicit models. arXiv preprint arXiv:2206.05564  (2022)

\bibitem{zheng2023dpmv3}
Zheng, K., Lu, C., Chen, J., Zhu, J.: Dpm-solver-v3: Improved diffusion ode solver with empirical model statistics. arXiv preprint arXiv:2310.13268  (2023)

\end{thebibliography}

\title{Supplementary Material of FRDiff} 


\authorrunning{J. So et al.}
\author{}

\institute{}
\renewcommand{\thefigure}{S\arabic{figure}}
\renewcommand{\thetable}{S\arabic{table}}
\maketitle

\section{Overview}

In this supplementary material, we provide a more detailed explanation of our implementation and additional experimental results. We include the following items:

\begin{itemize}
\item Detailed specification of $\mathcal{F}(\cdot)$ and skippable parts $\mathcal{S}(\cdot)$ for various diffusion models in Sec. \ref{sec:archi}
\item Proof of equivalence between output reusing and reduced NFE in Sec. \ref{sec:proof-of-jump}
\item Detailed implementation, configuration, and trained results of AutoFR in Sec. \ref{sec:autofr}
\item Quantitative results of Fig. 7 in the main paper in Sec. \ref{sec:quant}
\item Ablation study of feature reusing layer selection in  \ref{sec:layer-ablation}
\item Comparison with other feature-reusing methods in \ref{sec:other-fr}
\item Further measurements of skippable latency for other models and datasets in Sec. \ref{sec:latency}
\item Additional experiments on temporal similarity for other models and datasets in Sec. \ref{sec:temp}
\item Visual comparison of frequency response of reduced NFE and FR in Sec. \ref{sec:visual}
\item Additional qualitative results in Sec. \ref{sec:quali}
\item Experimental results on additional generation tasks, including image-to-video generation, super resolution, and image inpainting, in Sec. \ref{sec:task}

\item Discussion on potential negative impacts in Sec. \ref{sec:ethical}
\end{itemize}

\section{Detailed Model Architecture Specification}
\label{sec:archi}
\subsection{Diffusion U-Net}
Firstly, we introduce the overall architecture of the Diffusion U-Net, which currently stands as the most commonly utilized architecture in various diffusion models such as DDPM \cite{ho2020denoising}, LDM \cite{ldm}, SDXL \cite{sdxl}, and others. The Diffusion U-Net consists of two types of residual blocks: \texttt{ResNetBlock} and \texttt{SpatialTransformerBlock}. The specific structure of the \texttt{ResNetBlock} and its $\mathcal{S}(\cdot)$ is as follows:

\begin{equation}
\begin{aligned}
&\left.\begin{aligned}
\bold{x}_1 &\leftarrow  \texttt{GroupNorm}(\bold{x})\\ 
\bold{x}_1 &\leftarrow  \texttt{Conv}(\bold{x}_1)\\
\end{aligned}\right\}\hspace{0.5cm}\mathcal{S}(\cdot)\\
&\begin{aligned}
\bold{x}_1 &\leftarrow \bold{x}_1 + \texttt{MLP}(t)\\
\bold{x}_1 &\leftarrow  \texttt{GroupNorm}(\bold{x}_1)\\
\bold{x}_1 &\leftarrow  \texttt{Conv}(\bold{x}_1)\\
\bold{y}&\leftarrow \bold{x}_1 + \bold{x}\\
\end{aligned}\
\end{aligned}
\end{equation}
As shown, $\mathcal{S}(\cdot)$ takes roughly 50\% operations in its block. 
Next, the structure of the \texttt{SpatialTransformerBlock} is as follows:
\begin{equation}
\begin{aligned}
&\left.\begin{aligned}
\bold{x}_1 &\leftarrow  \texttt{GroupNorm}(\bold{x})\\ 
\bold{x}_1 &\leftarrow  \texttt{MLP}(\bold{x}_1)\\
\bold{x}_2 &\leftarrow  \texttt{LayerNorm}(\bold{x}_1)\\ 
\bold{x}_2 &\leftarrow  \texttt{SelfAttention}(\bold{x}_2) + \bold{x}_1\\
\bold{x}_3 &\leftarrow  \texttt{LayerNorm}(\bold{x}_2)\\ 
\bold{x}_3 &\leftarrow  \texttt{CrossAttention}(\bold{x}_3, c) + \bold{x}_2\\
\bold{x}_4 &\leftarrow  \texttt{LayerNorm}(\bold{x}_3)\\ 
\bold{x}_4 &\leftarrow  \texttt{MLP}(\bold{x}_4) + \bold{x}_3\\
\bold{x}_4 &\leftarrow  \texttt{MLP}(\bold{x}_4)
\end{aligned}\right\}\hspace{0.5cm}\mathcal{S}(\cdot)\\
&\begin{aligned}
\bold{y}&\leftarrow \bold{x}_4 + \bold{x}\\
\end{aligned}\
\end{aligned}
\end{equation}

Because the \texttt{SpatialTransformerBlock} does not incorporate time step information, we simply select $\mathcal{S}(\cdot)$ as the entire computation before the final residual operation.

\subsection{Diffusion Transformer \cite{dit}}

Next, we provide a detailed architecture specification for the DiT (Diffusion Transformer), which is a recently highlighted diffusion model architecture. Specifically, we utilize the adaLN-Zero version of the DiT architecture. Each DiT adaLN-Zero block is composed of two consecutive different residual blocks, self-attention, and feed-forward. The specification for the DiT self-attention block is as follows:

\begin{equation}
\begin{aligned}
&\left.\begin{aligned}
\bold{x}_1 &\leftarrow  \texttt{LayerNorm}(\bold{x})\\
\bold{x}_1 &\leftarrow  \gamma_1(t)*\cdot\bold{x}_1 + \beta_1(t)\\
\bold{x}_1 &\leftarrow  \texttt{SelfAttention}(\bold{x}_1) \\
\end{aligned}\right\}\hspace{0.5cm}\mathcal{S}(\cdot)\\
&\begin{aligned}
\bold{x}_1 &\leftarrow \alpha_1(t)*\bold{x}_1 + x
\end{aligned}\
\end{aligned}
\end{equation}
,where $\gamma(\cdot), \beta(\cdot), \alpha(\cdot)$ is MLP that predicts scaling, shift factor from time step information.
The specification for DiT-feed forward block is as follows :
\begin{equation}
\begin{aligned}
&\left.\begin{aligned}
\bold{x}_2 &\leftarrow  \texttt{LayerNorm}(\bold{x}_1)\\
\bold{x}_2 &\leftarrow  \gamma_2(t)*\cdot\bold{x}_2 + \beta_1(t)\\
\bold{x}_2 &\leftarrow  \texttt{MLP}(\bold{x}_2) \\
\end{aligned}\right\}\hspace{0.5cm}\mathcal{S}(\cdot)\\
&\begin{aligned}
\bold{y} &\leftarrow \alpha_2(t)*\bold{x}_2 + \bold{x}_1
\end{aligned}\
\end{aligned}
\end{equation}

While there are 3 types of time information($\alpha,\beta,\gamma$) injected into DiTBlock, we decided to only recompute $\alpha(\cdot)$ to achieve better acceleration.

For more information about entire architecture and tensor dimension, please refer to the original papers \cite{ho2020denoising, ldm, sdxl, dit} and the official codebases \footnote{https://github.com/CompVis/stable-diffusion} \footnote{ https://github.com/facebookresearch/DiT}.

\section{Proof of Equivalence between Output Reusing and reduced NFE}
\label{sec:proof-of-jump}
In this section, we provide a proof that reusing the entire output score of the model for each consecutive step, is identical to reducing the NFE (Number of Function Evaluations).

We provide proof of this statement in the case of DDIM \cite{ddim}. The reverse process of DDIM at time  $t$ is as follows:
\begin{equation}
    x_{t-1} = \frac{\sqrt{\Bar{\alpha}_{t-1}}}{\sqrt{\Bar{\alpha}_t}} (x_t - \sqrt{1-\Bar{\alpha}_{t}}\epsilon_{\theta}(x_t)) \\ + \sqrt{1-\Bar{\alpha}_{t-1}}\epsilon_\theta(x_t)
\label{eq:ddim-reverse}
\end{equation}
The model predicts the score of the data $\epsilon_{\theta}(x_t)$ at time t, and this score is used for denoising. Consider the case where the score obtained at time t ($\epsilon_{\theta}(x_t)$) is used for the next time $t-1$, as in Eq.~\ref{eq:ddim-reverse-reuse}. 

\begin{equation}
    x_{t-2} = \frac{\sqrt{\Bar{\alpha}_{t-2}}}{\sqrt{\Bar{\alpha}_{t-1}}} (x_{t-1} - \sqrt{1-\Bar{\alpha}_{t-1}}\epsilon_{\theta}(x_{t})) \\ + \sqrt{1-\Bar{\alpha}_{t-2}}\epsilon_\theta(x_t)
\label{eq:ddim-reverse-reuse}
\end{equation}
\\
Then, combining Eq.~\ref{eq:ddim-reverse} and Eq.~\ref{eq:ddim-reverse-reuse}, it can be expressed as:
\begin{align}
    x_{t-2} = \frac{\sqrt{\Bar{\alpha}_{t-2}}}{\sqrt{\Bar{\alpha}_{t-1}}} (\frac{\sqrt{\Bar{\alpha}_{t-1}}}{\sqrt{\Bar{\alpha}_t}} (x_t - \sqrt{1-\Bar{\alpha}_{t}}\epsilon_{\theta}(x_t))& \nonumber + \sqrt{1-\Bar{\alpha}_{t-2}}\epsilon_\theta(x_t) 
\label{eq:proof-jump-2}
\end{align}
\\
Finally, the above equation can be represented as:
\begin{align}
    x_{t-2} = \frac{\sqrt{\Bar{\alpha}_{t-2}}}{\sqrt{\Bar{\alpha}_{t}}} (x_t - \sqrt{1-\Bar{\alpha}_{t}}\epsilon_{\theta}(x_t)) + \sqrt{1-\Bar{\alpha}_{t-2}}\epsilon_\theta(x_t)
\end{align}
This result corresponds to the reverse process over an interval of 2 at time $t$ in DDIM. Therefore, consistently using the output score of the model at time $t$  aligns with the goal of reducing the NFE.

\section{Details on AutoFR}
\label{sec:autofr}
\subsection{Cost Loss}

To regulate the number of Feature Reuses in AutoFR, we introduce an additional cost loss function alongside the MSE loss, as elucidated in Eq. 13 in the main paper. This cost function is computed as follows:

\begin{equation}
\mathcal{L}_{cost}(\theta) = \sum_{t}^{N} \text{ReLU}(\text{sigmoid}(\theta_t) - 1/2)
\end{equation}

Here, The ReLU function is employed to regularize values greater than 0. Since $\theta_t = 0$ denotes feature reuse at time step $t$, we can effectively regulate the computational load during the denoising process with Feature Reuse.

This additional cost loss serves to penalize excessive feature reuse, thereby promoting a balanced utilization of computational resources throughout the process. By incorporating this regularization term, AutoFR attains a more controlled approach to feature reuse, optimizing performance while managing computational overhead.

\subsection{Training Recipe}
In Table \ref{tab:hyp_autofr}, we present the hyperparameters and experimental configurations of AutoFR. The hyperparameter $\lambda$ will be discussed in the next section. We found that a small number of training iterations are sufficient for convergence, so we decided to utilize 100-200 training iterations. All experiments were conducted using GPU servers equipped with 8 NVIDIA RTX 4090 GPUs.

\begin{table}[ht]
\centering
\begin{tabular}{c|cccccc}
\toprule
Model &  lr & optimizer & $\beta_1$ & $\beta_2$ & iteration \\
\midrule
SD  & 5e-2 & Adam & 0.9 & 0.999 & 150 \\
DiT  & 1e-3 & Adam & 0.9 & 0.999 & 100 \\
\bottomrule
\end{tabular}
\caption{Hyperparameteres of AutoFR}
\label{tab:hyp_autofr}
\end{table}

\section{Quantitative Results of Fig 7}
\label{sec:quant}

\subsection{Pareto Points}
In Table \ref{tab:stable_diffusion}, \ref{tab:dit_256}, we provide the quantitative results of Fig 7 of main paper. Also, we provide additional metrics measurement such as sFID, Recall, Precision for better comparison in DiT-256.
\begin{table}[ht]
\centering
\begin{minipage}{0.35\textwidth}
\centering
\begin{tabular}{cc|c|cccc}
\toprule
\multicolumn{4}{c}{Stable Diffusion}\\
\toprule
\; NFE & \; M \; & \; Latency \;  & \; FID \; \\
\midrule
\multirow{3}{*}{50} & 1 & 7.542 & 4.84  \\
& 2 & 4.184 & 6.40  \\
& 3 & 3.111 & 7.60  \\
\midrule
\multirow{3}{*}{40} & 1 & 6.051 & 5.79  \\
& 2 & 3.338 & 7.24 \\
& 3 & 2.548 & 8.66  \\
\midrule
\multirow{3}{*}{30} & 1 & 4.652 & 6.49  \\
& 2 & 2.491 & 7.83  \\
& 3 & 1.833 & 9.66  \\
\midrule
\multirow{3}{*}{25} & 1 & 3.862 & 7.66  \\
& 2 & 2.092 & 9.25  \\
& 3 & 1.547 & 11.28  \\
\midrule
\multirow{2}{*}{20} & 1 & 3.013 & 8.45  \\
& 2 & 1.918 & 10.22  \\
\midrule
\multirow{1}{*}{10} & 1 & 1.523 & 12.44  \\
\bottomrule
\end{tabular}
\caption{Stable Diffusion}
\label{tab:stable_diffusion}
\end{minipage}
\hfill
\begin{minipage}{0.6\textwidth}
\centering

\begin{tabular}{cc|c|cccc}
\toprule
\multicolumn{7}{c}{DiT-256}\\
\toprule
\; NFE & \; M \; & \; Latency \; & \; FID & \; sFID & \; Recall & \; Precision\\
\midrule
\multirow{3}{*}{50} & 1 & 1.270 & \; 13.34 & \; 19.0  & \; 0.748 & 0.665 \\
                    & 2 & 0.775 & \; 13.48 & \; 18.06 & \; 0.736 & 0.669 \\
                    & 3 & 0.611 & \; 14.82 & \; 18.15 & \; 0.729 & 0.655 \\
\midrule
\multirow{3}{*}{40} & 1 & 1.017 & \; 13.76 & \; 18.91 & \; 0.747 & 0.662 \\
                    & 2 & 0.619 & \; 14.58 & \; 18.14 & \; 0.734 & 0.657 \\
                    & 3 & 0.496 & \; 16.87 & \; 19.23 & \; 0.716 & 0.644 \\
\midrule
\multirow{3}{*}{30} & 1 & 0.763 & \; 14.76 & \; 18.95 & \; 0.743 & 0.66 \\
                    & 2 & 0.463 & \; 16.81 & \; 18.65 & \; 0.729 & 0.634 \\
                    & 3 & 0.360 & \; 21.71 & \; 21.13 & \; 0.709 & 0.597 \\
\midrule
\multirow{2}{*}{25} & 1 & 0.644 & \; 15.73 & \; 18.99 & \; 0.744 & 0.648 \\
                    & 2 & 0.396 & \; 19.31 & \; 19.51 & \; 0.716 & 0.615 \\
\midrule
\multirow{2}{*}{20} & 1 & 0.515 & \; 17.69 & \; 19.31 & \; 0.738 & 0.635 \\
                    & 2 & 0.396 & \; 19.31 & \; 19.51 & \; 0.715 & 0.615 \\
\midrule
\multirow{1}{*}{10} & 1 & 0.257 & \; 37.53 & \; 25.84 & \; 0.699 & 0.491 \\
\bottomrule
\end{tabular}
\caption{DiT-256}
\label{tab:dit_256}
\end{minipage}
\end{table}

\subsection{AutoFR results}
In this section, we provide the \textit{keyframe sets} searched by AutoFR and corresponding $\lambda$ that denoted in Fig 7 of main paper. 

The searched keyframe sets in Stable Diffusion (Fig. 7 (a) of main paper) are as follows :
\begin{itemize}
    \item $\mathcal{K} = \{1,2,3,4,5,6,7,8,9,11,12,13,14,15,16,17,21,23,26,27,28,33,36,38,\},\\ \lambda=1e-4, N=40$
    \item $\mathcal{K} = \{1,2,3,4,5,6,7,8,9,10,12,13,14,17,18,22,26,30,32,34\},\\ \lambda=1e-4, N = 35$
    \item $\mathcal{K} = \{1,2,3,4,5,6,7,8,9,12,13,18,21,22,23,24,27,31,33\},\\ \lambda=1e-3, N=35$

\end{itemize}

The searched keyframe set in DiT (Fig. 7 (b) of main paper) are as follows :

\begin{itemize}
    \item $\mathcal{K} = \{1,4,5,6,7,11,12,13,14,15,16,18,19,20,22,23,24,25,26,28,\\ 
    29,31,34,37\}, N=40, \lambda=0.01$
\end{itemize}

As can be observed in the results, the searched keyframe set tends to jump more frequently during the very initial denosing stage and later denoising stage. This is because if there are many jumps during the initial denoising step, the initial score estimation becomes inaccurate, leading to more accumulated errors that adversely affect the final generated result.

\section{Ablation Study of Reusing Layer Selection}
\label{sec:layer-ablation}
In this section, we conducted an ablation study to investigate the impact of reusing only the ResBlocks/Transformer and Encoder/Decoder blocks in U-Net. The FID/Latency was measured using SDv1.4 on the MS-COCO dataset. As depicted in Table \ref{tab:unet_ablation}, reusing both blocks results in the lowest latency with nearly identical FID scores.  This outcome is likely due to the negative impact of mixing time step information between skipped and non-skipped sections when layers are partially skipped. Hence, we decided to reuse all layers together in out FRDiff.
\begin{table}[h]
    \centering
\begin{tabular}{ccc|cc}
        \toprule
        Reuse Layer  & Step & M & FID & Latency \\
        \midrule
        ResBlock & 50 & 2 & 6.351 & 5.639\\ 
        Transformer & 50 & 2 & 6.461 & 5.161\\ 
        \midrule
        Encoder & 50 & 2 & 6.376 & 5.572\\ 
        Decoder & 50 & 2 & 6.452 & 5.189\\ 
        \midrule
        \textbf{Both(FRDiff)} & 50 & 2 & 6.407 & \textbf{4.183}\\
        \bottomrule
      \end{tabular}
          \caption{Ablation study of Layer Selection. Reusing both Layer shows smallest latency with nearly identical FID.}
    \label{tab:unet_ablation}
\end{table}

\section{Comparison with Other Feature-Reusing Methods}
\label{sec:other-fr}
In this section, we compare the performance of our FRDiff method with other recently released feature reusing-based methods. Specifically, we compared FID / Latency on Stable Diffusion V1.4 with MS-COCO dataset using DeepCache\cite{deepcache}, Faster Diffusion\cite{fasterdiffusion}. As shown in Fig. \ref{fig:method_comparison}, FRDiff demonstrates
the best FID-latency trade-off. This is because while naive feature reusing damages low-frequency components, our FRDiff preserves both low- and high-frequency components through score mixing. Moreover, our AutoFR automatically finds the best feature reusing policy for diffusion models.

\begin{figure}[h!]
    \centering
        \includegraphics[width=0.5\columnwidth]{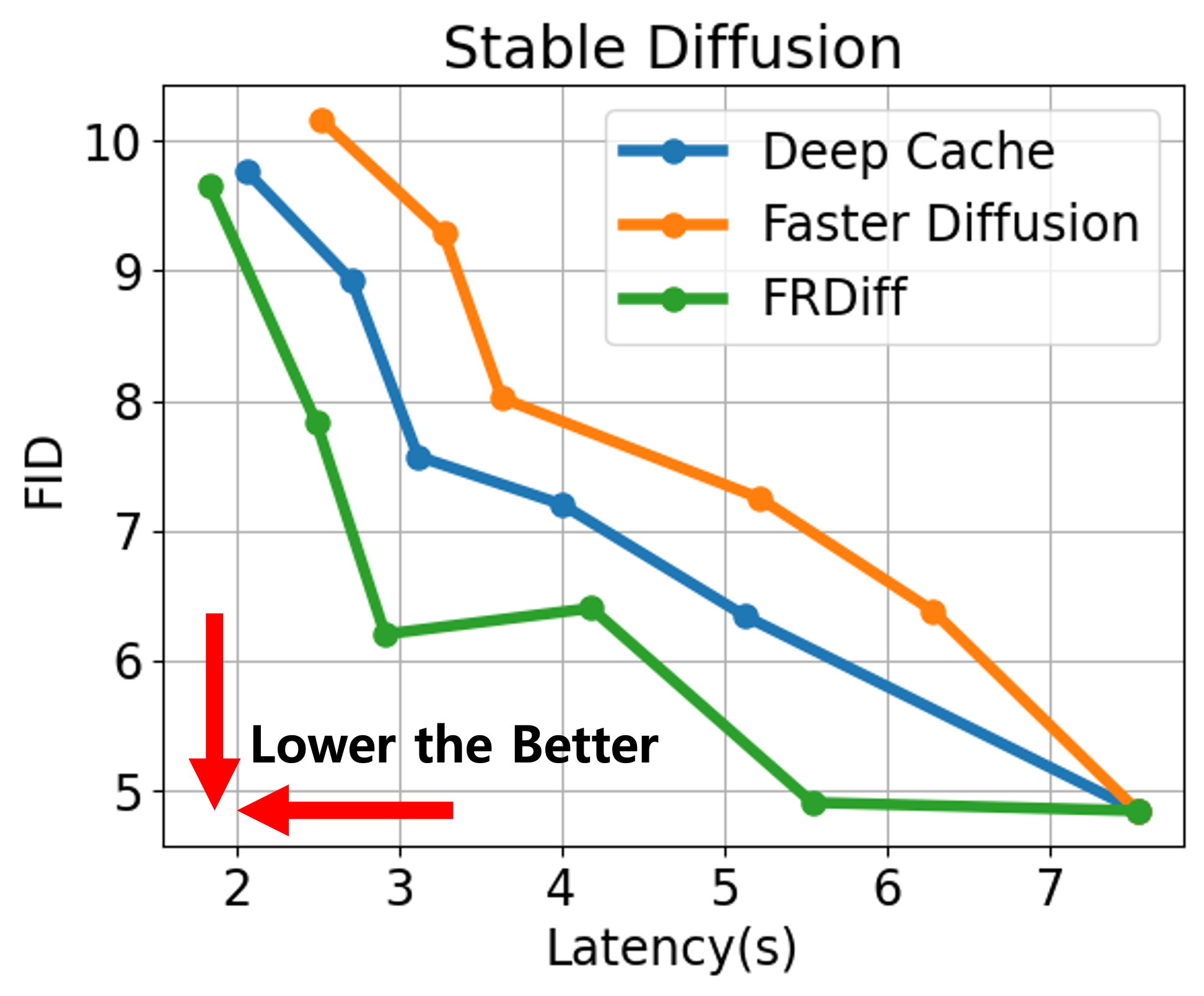}
    \caption{Comparison with other Feature Reusing methods}
    \label{fig:method_comparison}
\end{figure}

\section{Additional Measurement of Skippable Latency}
\label{sec:latency}
In Fig. \ref{fig:latency-other}, we present additional measurement of skippable latency in various diffusion models, such as Pixel-Space \cite{ho2020denoising}, LDM \cite{ldm}, Stable Diffusion \cite{ldm}, Stable Video Diffusion. As shown in figure, because the portion of skippable latency is large, we can achieve sufficient acceleration effect with FRDiff in various types of diffusion models.

\section{Additional Temporal Similarity Visualization}
\label{sec:temp}
In Fig \ref{fig:sim1}, \ref{fig:sim2}, we provide additional feature map visualization experiment of diffusion model in more layers and timesteps. As shown in figure, the temporal similarity of diffusion model is very high regardless of layer, timestep.

\section{Visual Comparison of Frequency Response}
\label{sec:visual}

In Fig. \ref{fig:method_comparison}, we present the original image (a) generated with DDIM over 50 steps on LDM-4 CelebA-HQ, FR with an interval of 10 (b), and output reusing with the same interval (or NFE=5) (c).  \textbf{Please note that we intentionally use a large interval to easily visualize the generation behavior of FR and output reusing.} In comparison to the original image (a), FR (b) effectively preserves details like hair texture but exhibits differences in color. Conversely, Jump (c) maintains color well but has a blurry image and struggles to preserve details. The Mix (d) image 
preserve relatively more frequency components than (b), (c).

\begin{figure}[t]
    \centering
        \includegraphics[width=0.7\columnwidth]{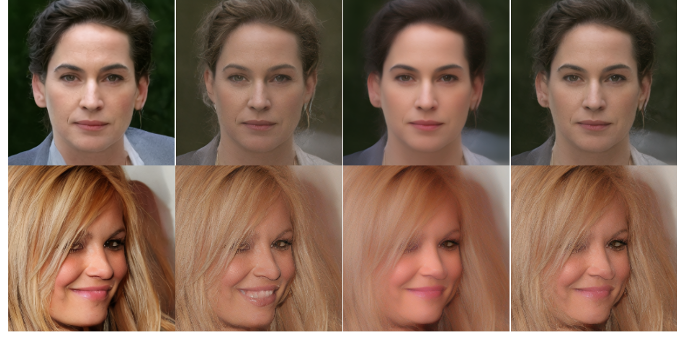}
    \caption{Generated images of LDM-CelebA-HQ \cite{celeba-hq} using various reusing schemes with DDIM 50 step interval 10. While reduced NFE (or output reusing) (c) shows blurry image, FR (b) effectively preserve details(e.g hair texture). However, (b) tends to compromise low-frequency components such as colors.
    Meanwhile, our proposed score mixing (d) preserves both low and high-frequency components well}
    \label{fig:method_comparison}
\end{figure}

\section{Additional Qualitative Results}
\label{sec:quali}
\begin{figure*}[t!]
    \centering
    \includegraphics[width=0.99\textwidth]{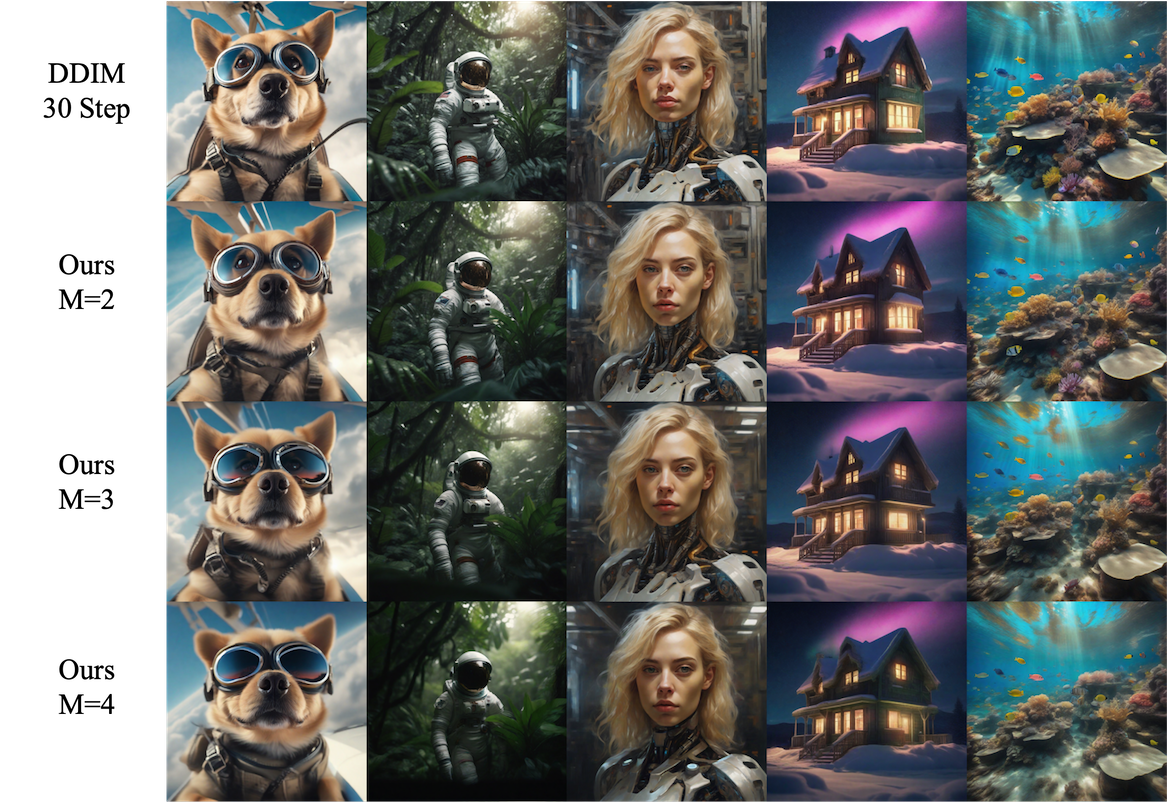}
    \caption{Additional generation results with FRDiff in SDXL}
    \label{fig:laasd1}
\end{figure*}

\begin{figure*}[t!]
    \centering
    \includegraphics[width=0.99\textwidth]{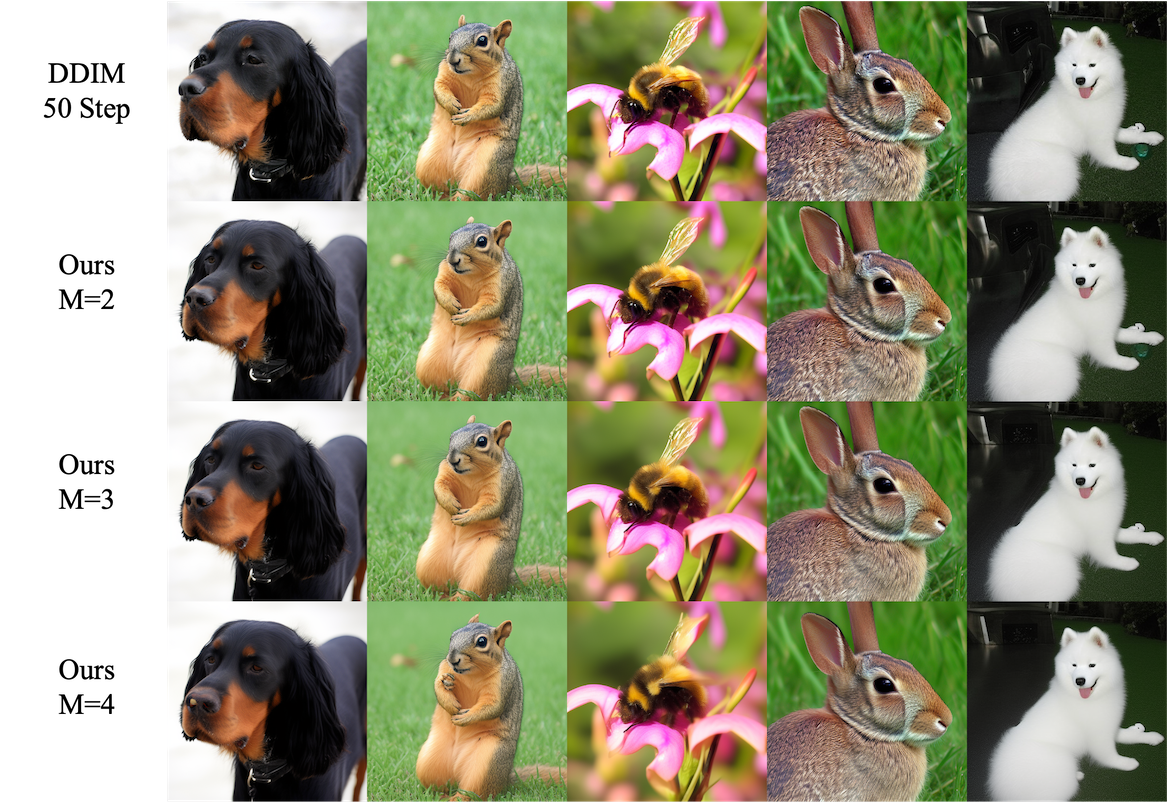}
    
    \caption{Additional generation results with FRDiff in DiT-512}
    \label{fig:laasd2}
\end{figure*}

\begin{figure*}[t!]
    \centering
    \includegraphics[width=0.95\textwidth]{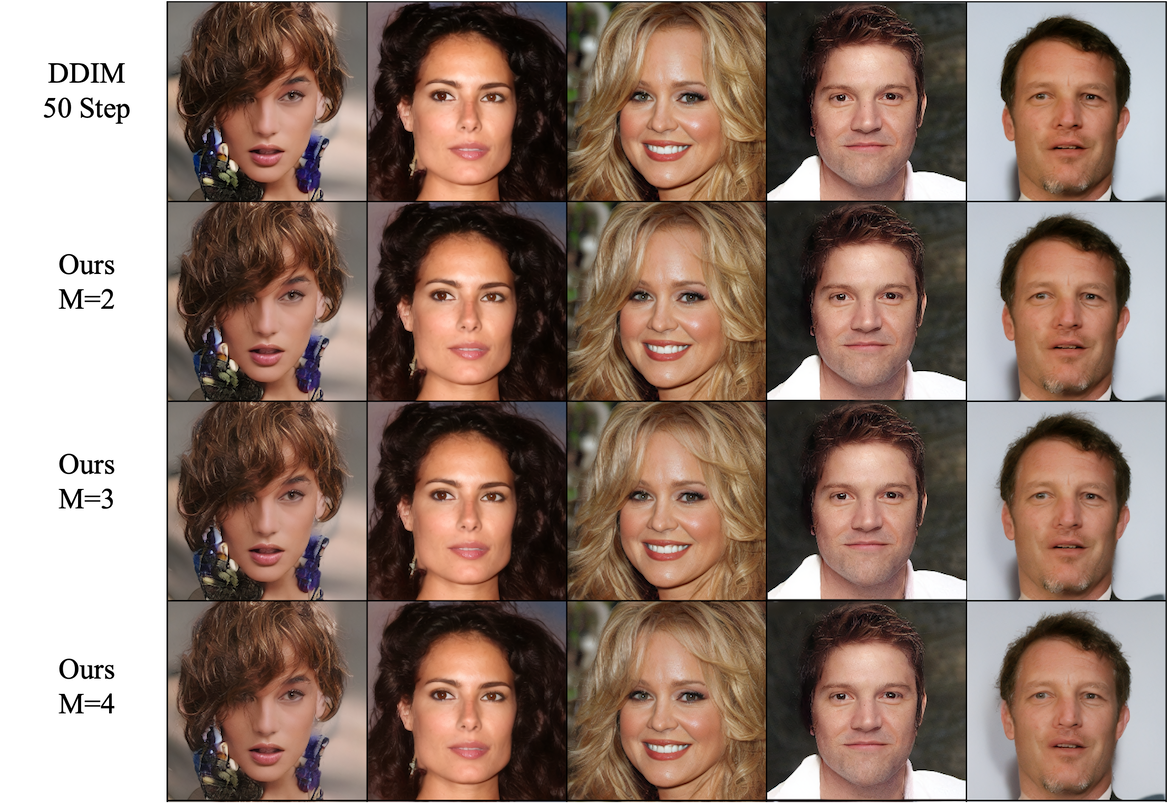}
    
    \caption{Additional generation results with FRDiff in LDM-4 (CelebA-HQ)}
    \label{fig:laasd3}
\end{figure*}

\begin{figure*}[t!]
    \centering
    \includegraphics[width=0.95\textwidth]{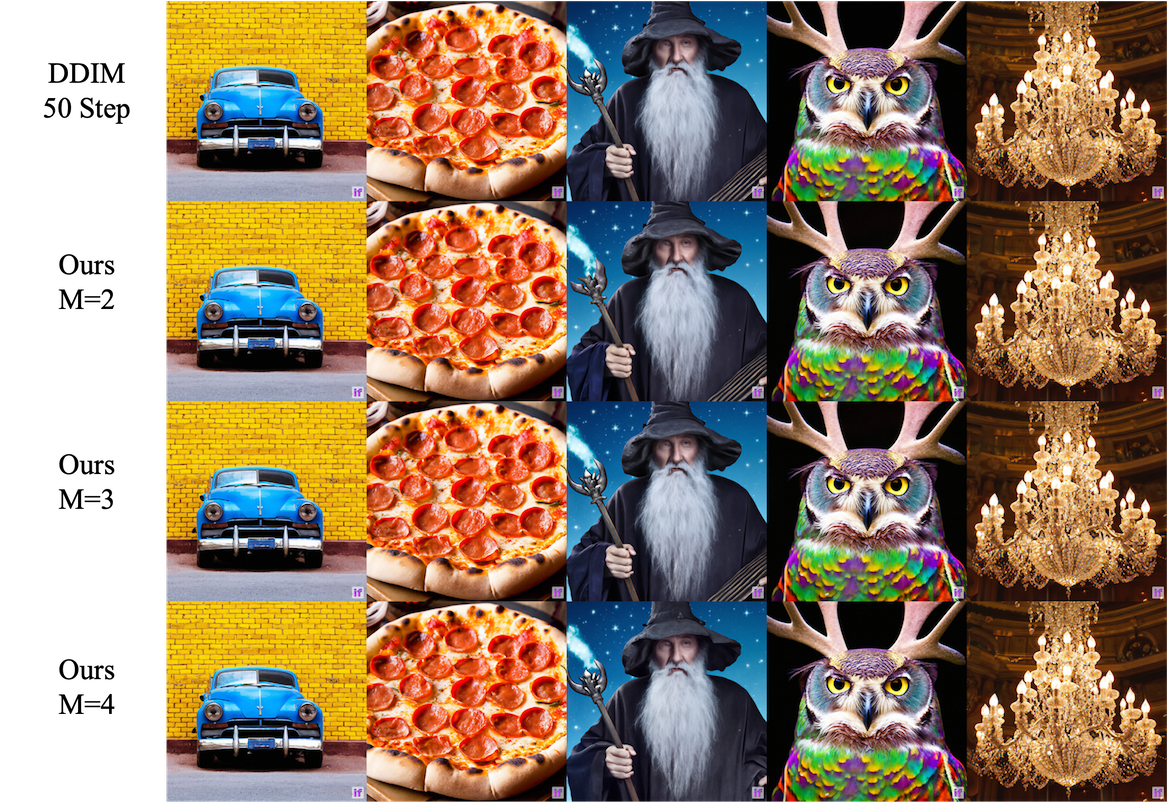}
    
    \caption{Additional generation results with FRDiff in DeepFloyd-IF. We only apply FRDiff to 3rd stage of sampling pipeline in DeepFloyd-IF.}
    \label{fig:laasd4}
\end{figure*}

\begin{figure*}[t!]
    \centering
    
    \begin{subfigure}[b]{0.45\columnwidth}
        \includegraphics[width=\textwidth]{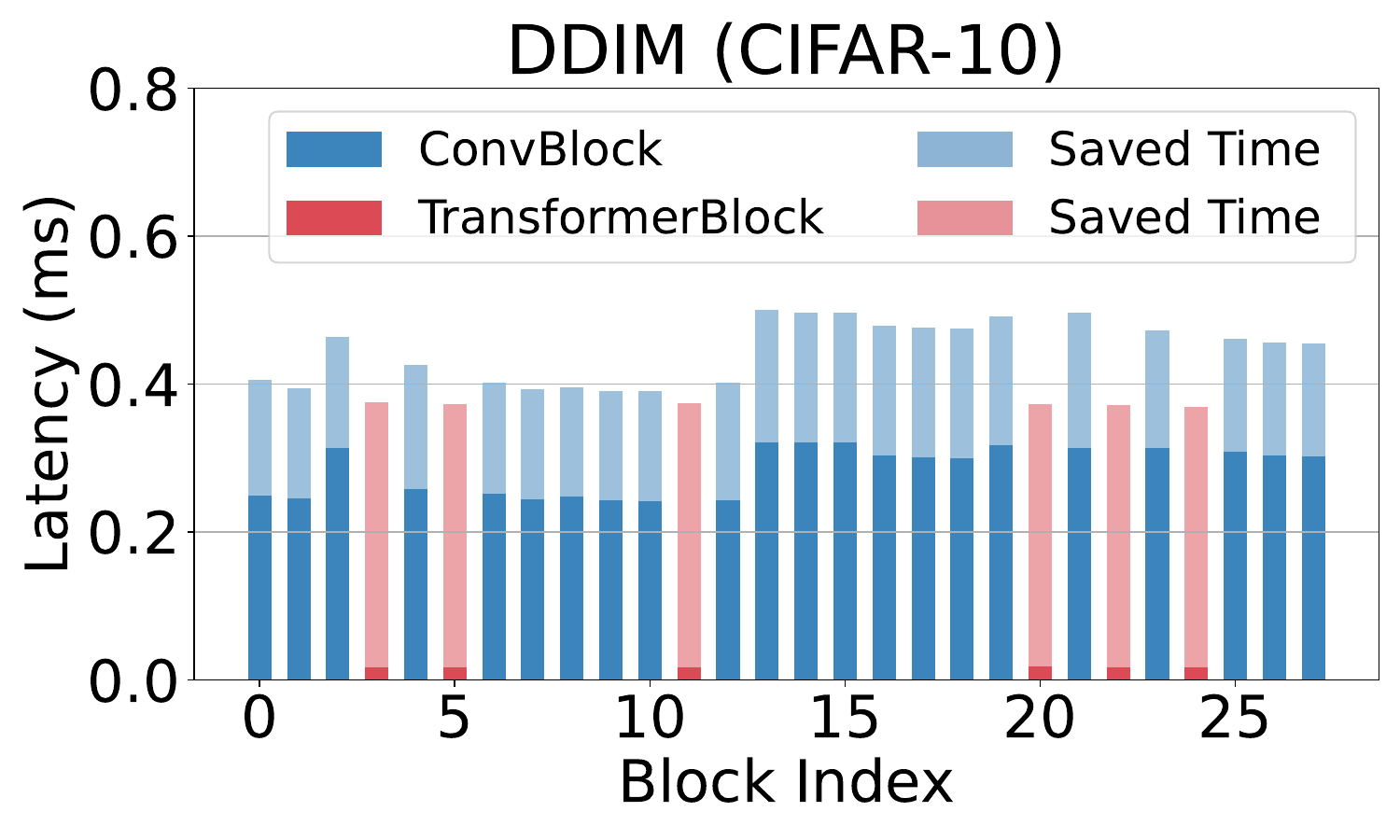}
        \caption{DDIM (CIFAR-10)}
    \label{fig:latency-cifar}
    \end{subfigure}
    \begin{subfigure}[b]{0.45\columnwidth}
        \includegraphics[width=\textwidth]{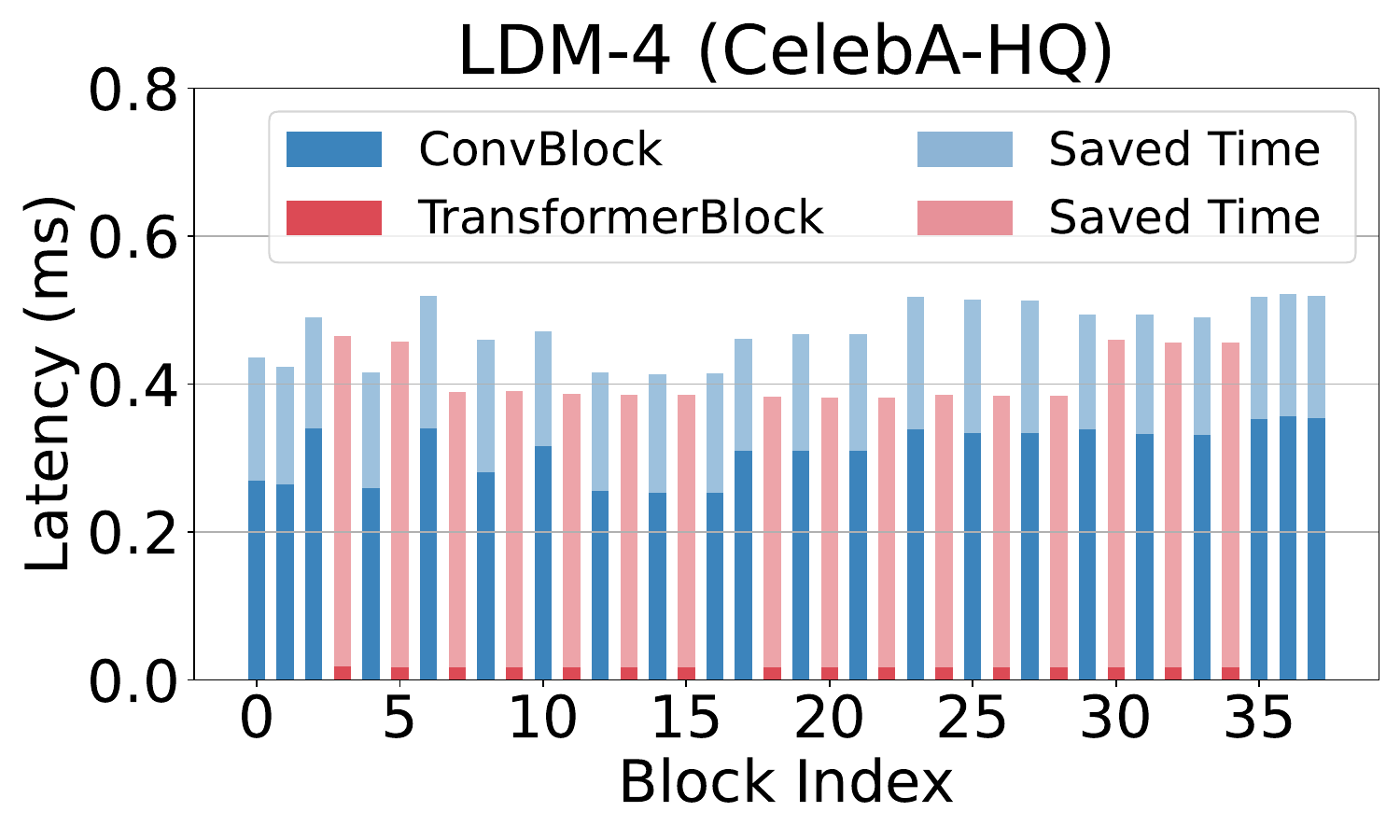}
        \caption{LDM-4 (CelebA-HQ)}
    \label{fig:latency-ldm}
    \end{subfigure}
    \begin{subfigure}[b]{0.45\columnwidth}
        \includegraphics[width=1.0\textwidth]{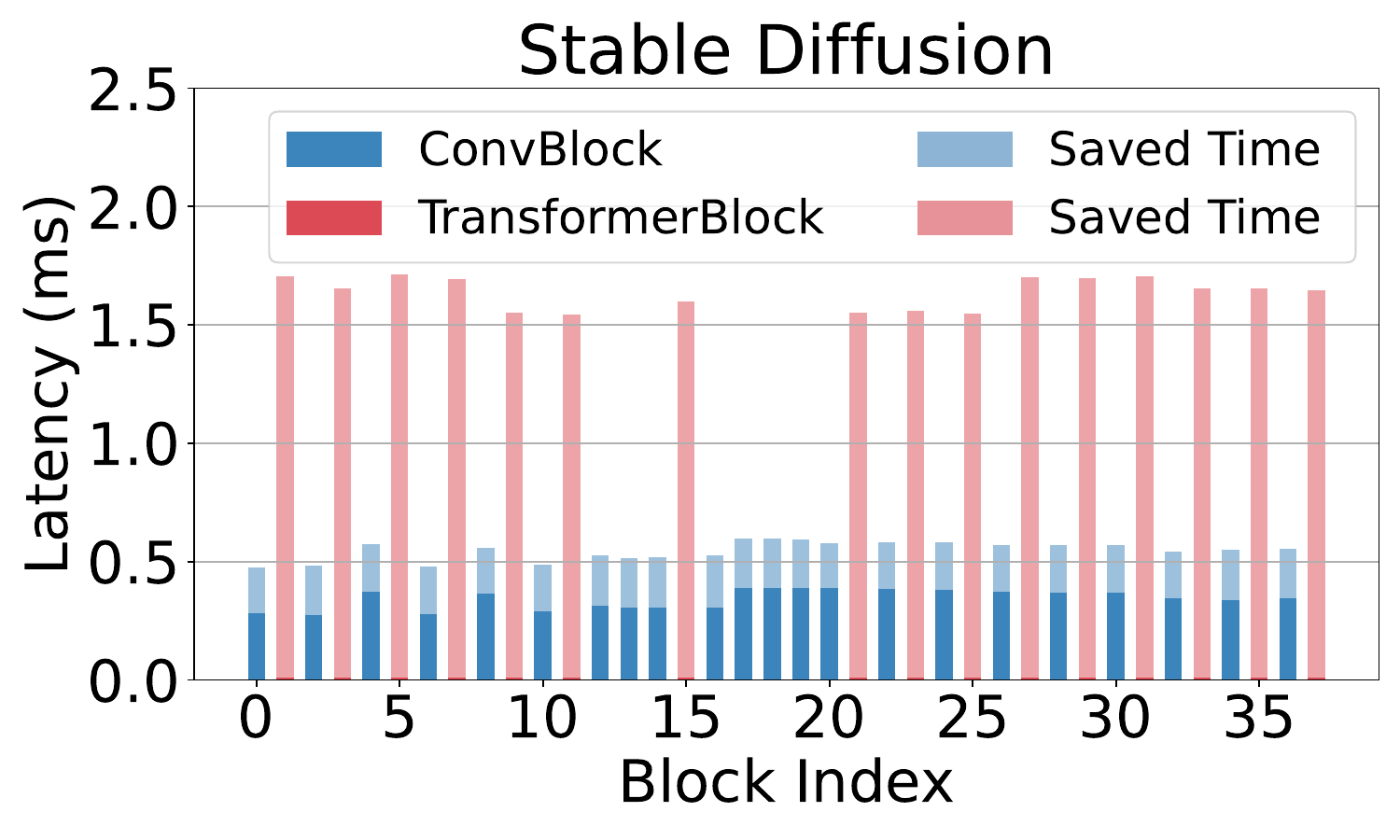}
        \caption{Stable Diffusion}
    \label{fig:latency-sd}
    \end{subfigure}
    \begin{subfigure}[b]{0.45\columnwidth}
        \includegraphics[width=1.0\textwidth]{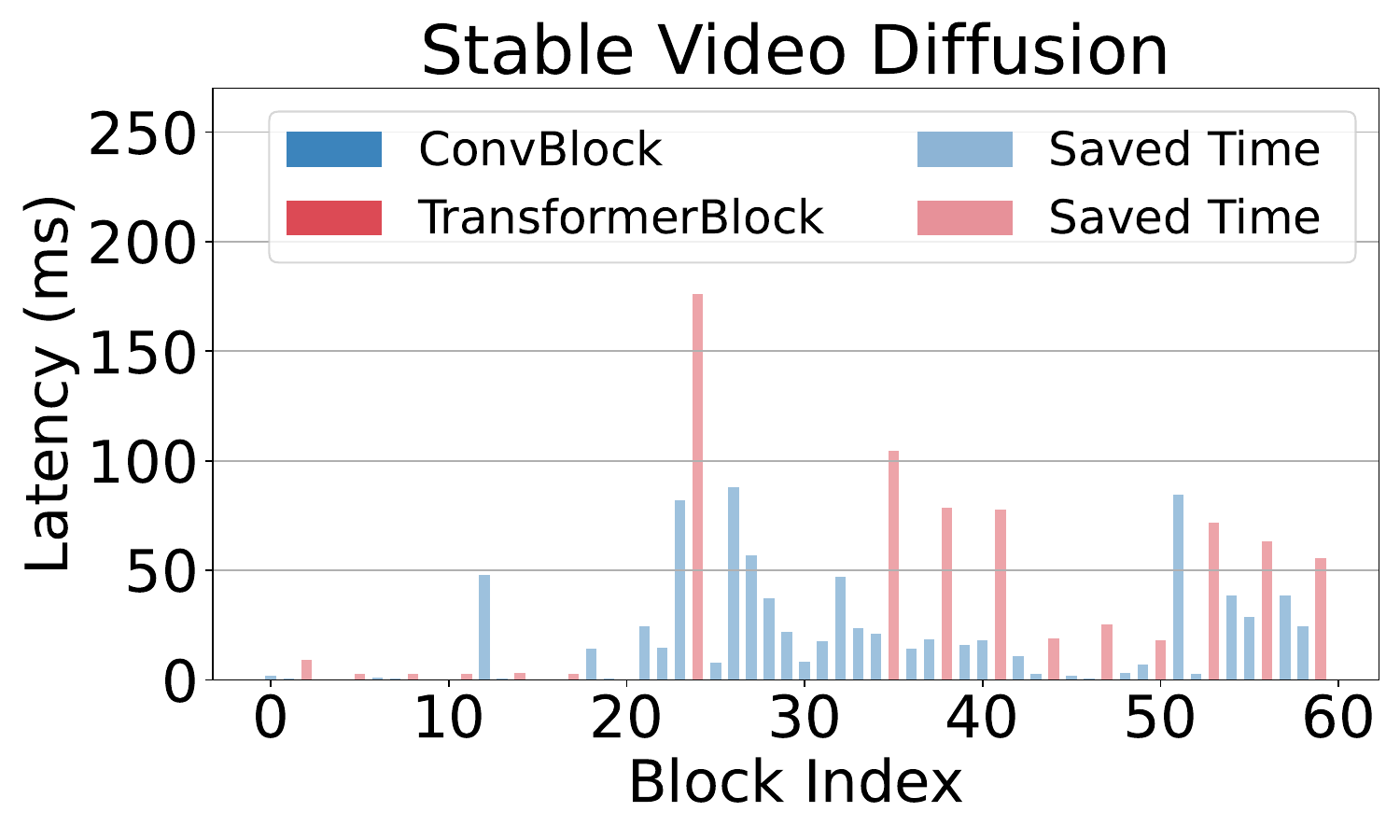}
        \caption{Stable Video Diffusion}
    \label{fig:latency-svd}
    \end{subfigure}
    
    \caption{Skippable latency with Feature Reuse}
    \label{fig:latency-other}
\end{figure*}

\begin{figure}
    \centering    
    \begin{subfigure}[b]{0.99\columnwidth}
        \includegraphics[width=\textwidth]{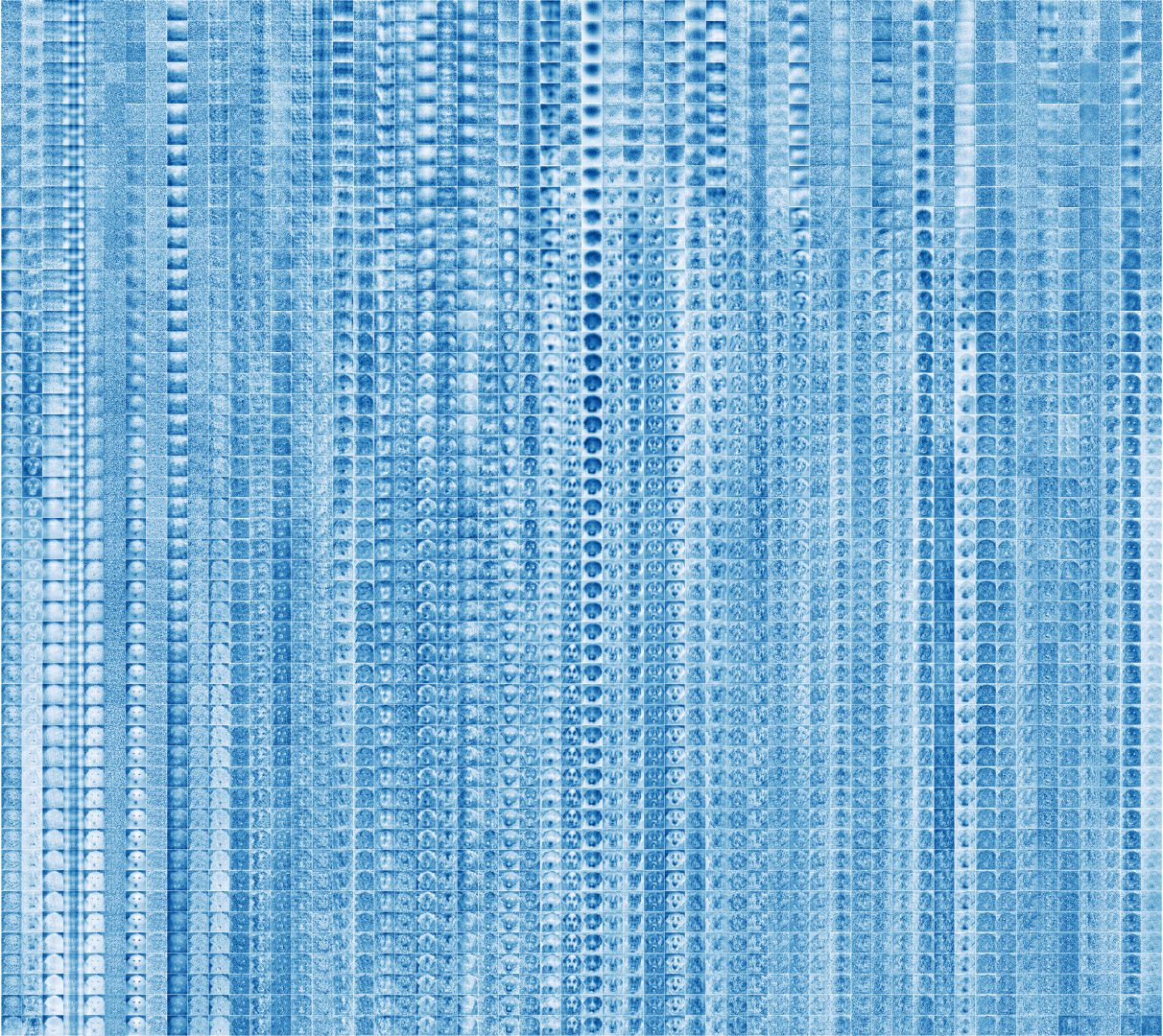}
    \end{subfigure}
    \caption{The feature map visualization results of DiT. we depicts the channel averaged values of each layer's feature map within denoising time step. \textit{Best viewed zoomed in}. The x-axis represent different layers and y-axis represent time step. }
    \label{fig:sim1}
\end{figure}

\begin{figure}[t]
    \centering
    \begin{subfigure}[b]{0.80\columnwidth}
        \includegraphics[width=\textwidth]{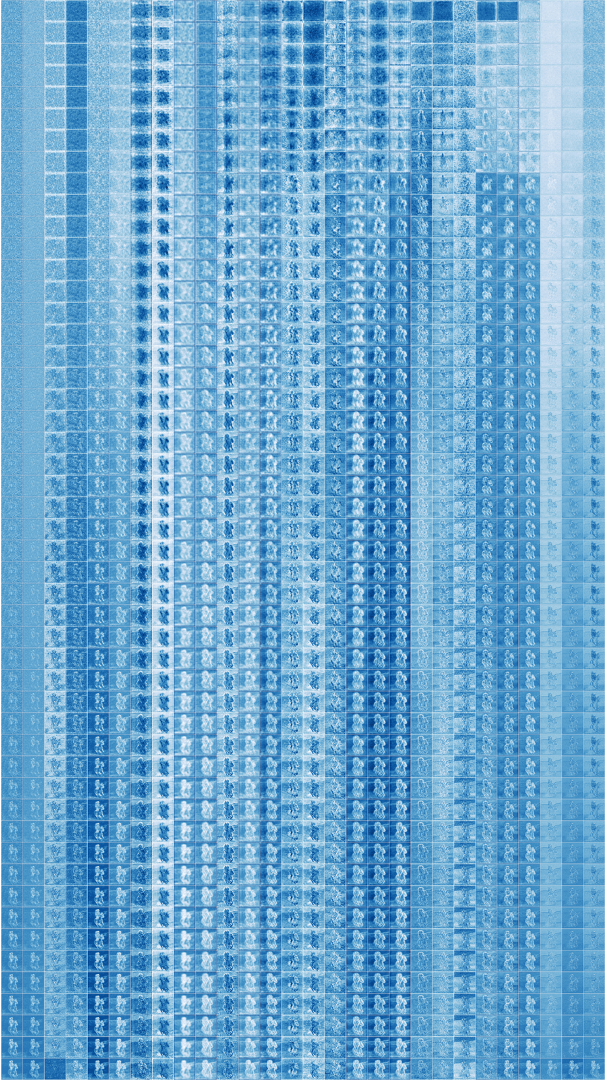}
                
    \end{subfigure}
    \caption{The feature map visualization results of SDXL. we depicts the channel averaged values of each layer's feature map within denoising time step. \textit{Best viewed zoomed in}. The x-axis represent different layers and y-axis represent time step. }
    \label{fig:sim2}
\end{figure}

In Fig. \ref{fig:laasd1} - \ref{fig:laasd4}, we provide additional qualitative results of our method. As shown in figure, our method consistently shows good generation quality in various instances or prompts regardless of types of model. 

\section{Additional Tasks Experiments}
\label{sec:task}
\subsection{Super Resolution}

\begin{figure*}
    \centering
    \begin{subfigure}[b]{0.49\columnwidth}
                \includegraphics[width=\textwidth]{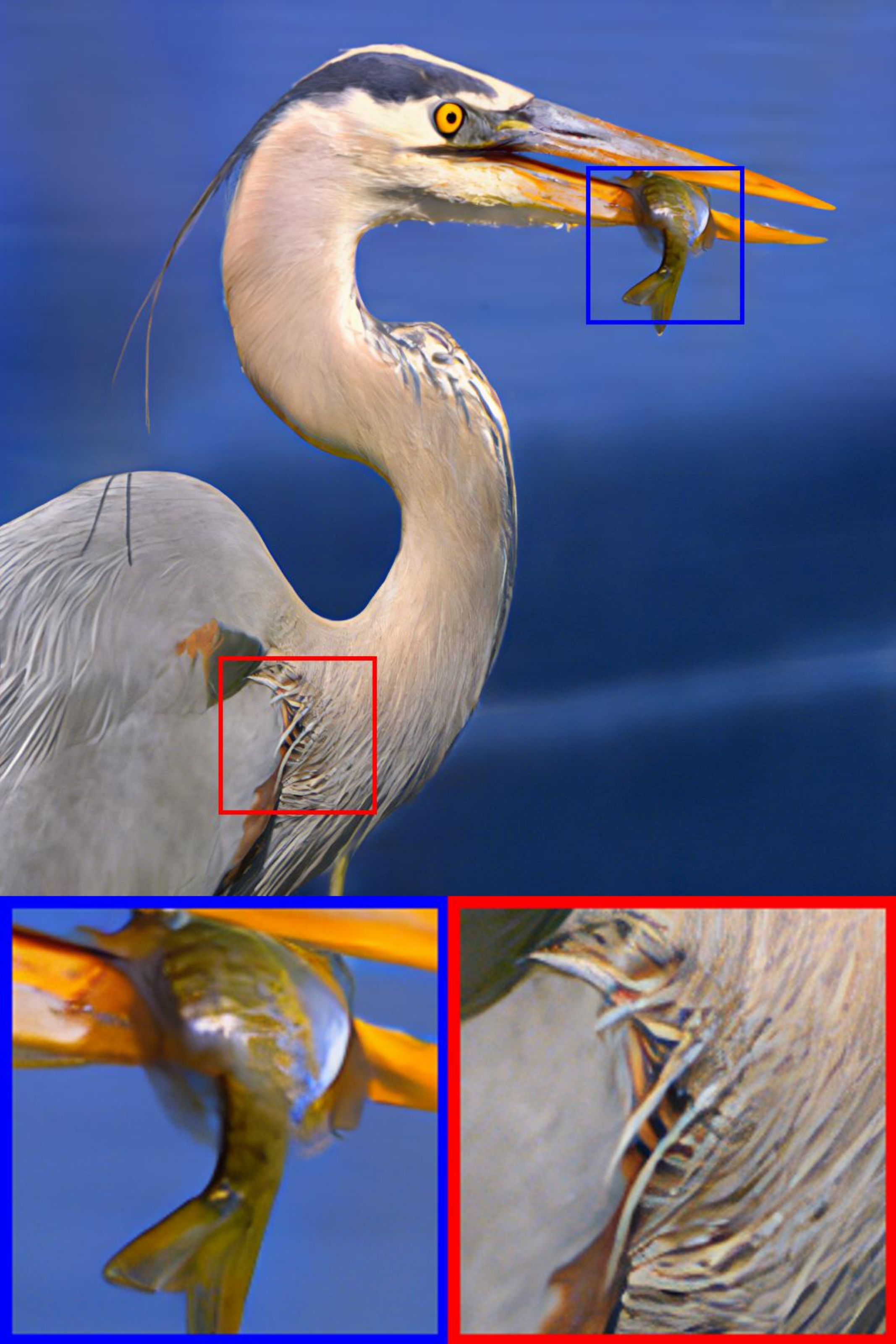}
    \end{subfigure}
    \begin{subfigure}[b]{0.49\columnwidth}
                \includegraphics[width=\textwidth]{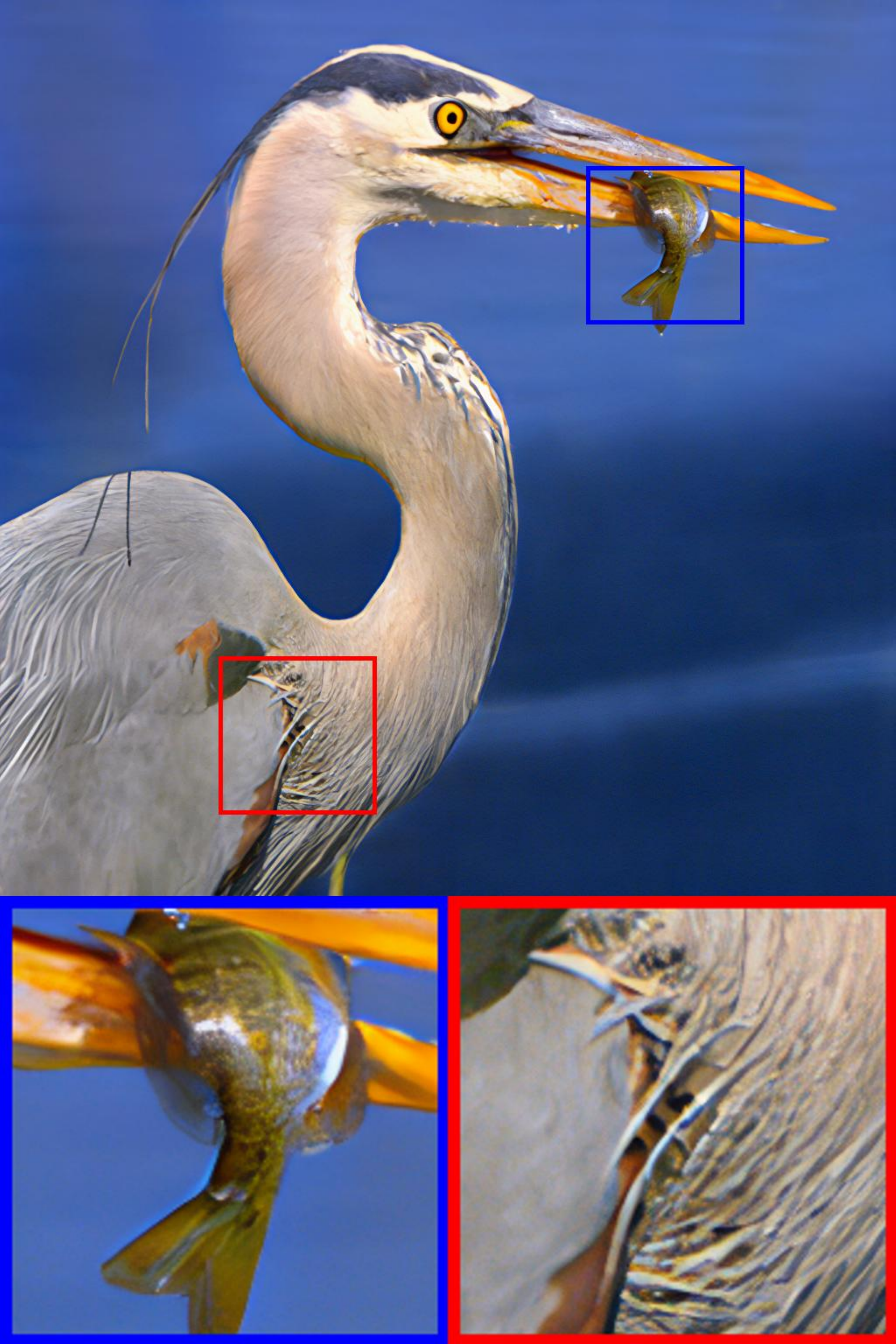}
    \end{subfigure} 

    \centering
    \begin{subfigure}[b]{0.49\columnwidth}
                \includegraphics[width=\textwidth]{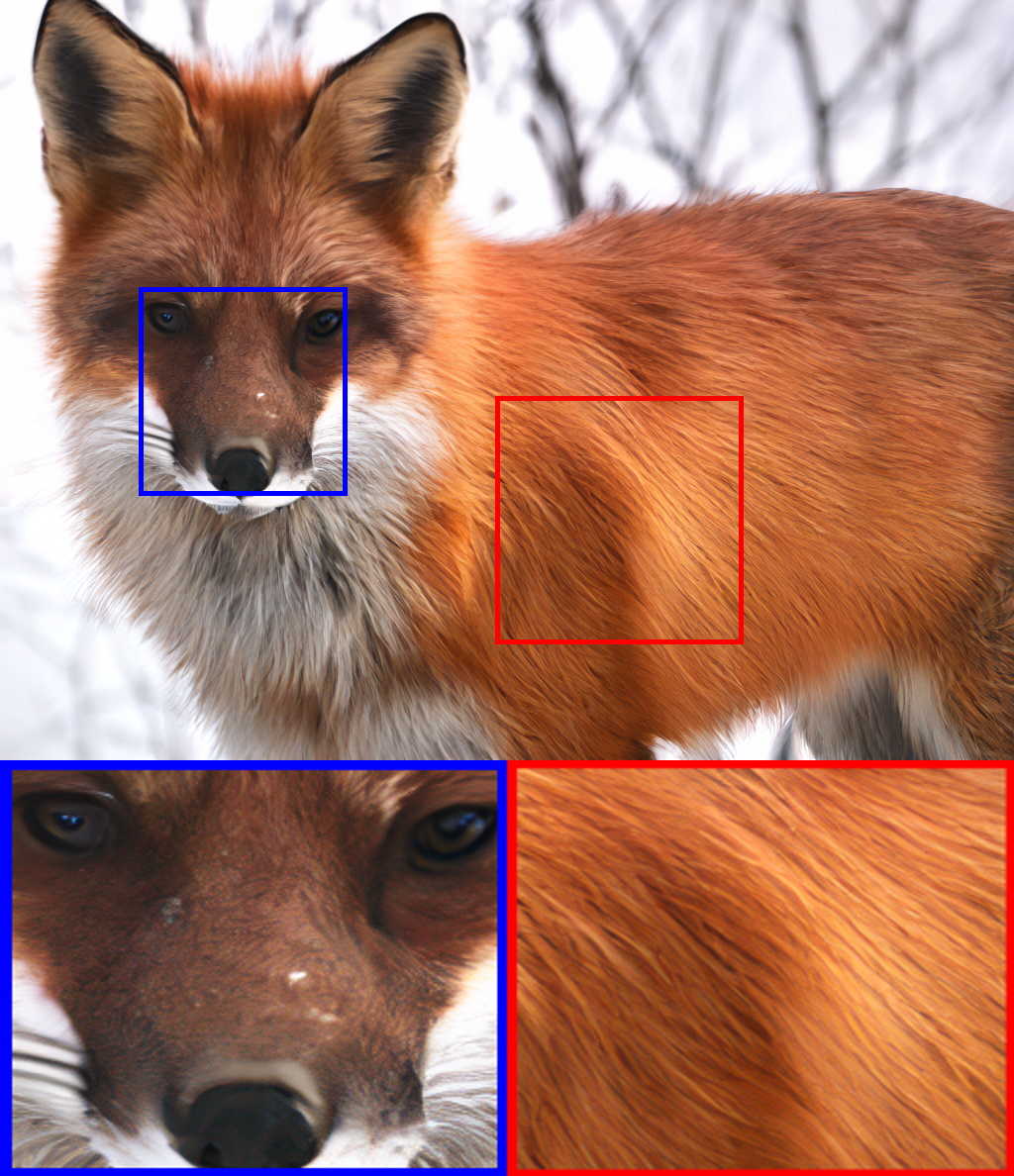}
                \caption{DDIM 30 Step}
    \end{subfigure}
    \begin{subfigure}[b]{0.49\columnwidth}
                \includegraphics[width=\textwidth]{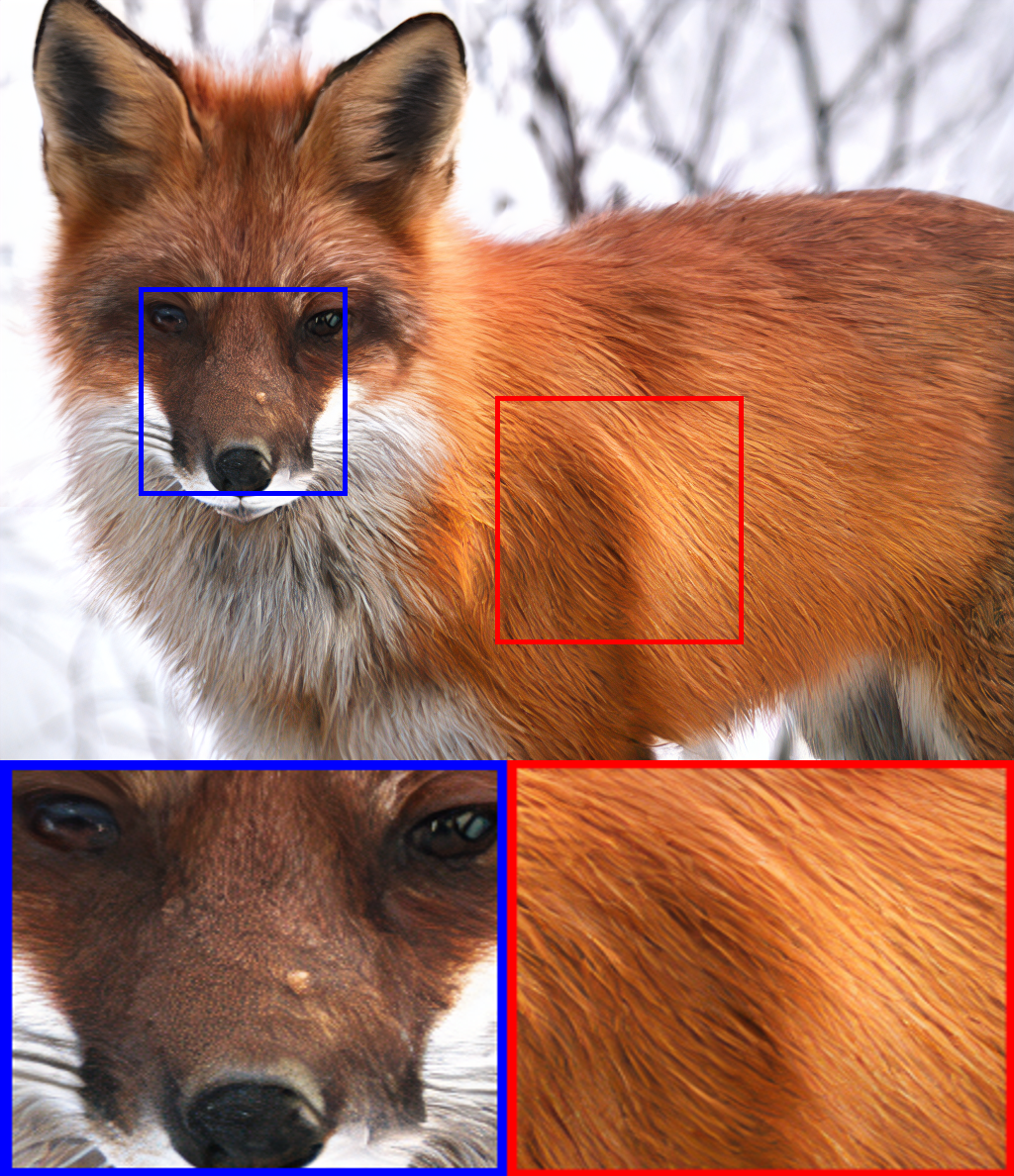}
                \caption{Ours}
    \end{subfigure}   
    \caption{LDM-SR super-resolution 4x upscaling result. In the regions highlighted by the blue and red boxes, Ours (DDIM 50 Step interval 3) (c) synthesizes more detailed textures effectively when compared to DDIM (b) with the same latency budget.}
    \label{fig:GRID-SR}
\end{figure*}

To assess the effectiveness of our model in task-specific applications, we performed image synthesis, upscaling a 256x192 resolution image to 1024x768 resolution using LDM-SR. In Fig.~\ref{fig:GRID-SR}, we compare our method (b) with the existing DDIM \cite{ddim} sampling (a) with the same latency budget, 7.64s. As depicted in the figure, DDIM (a) recovers some details but exhibits slightly blurred image texture. In contrast, our method (b) preserves better details than DDIM (a) and includes higher-quality features. This demonstrates that our method can generate higher-quality and more detailed images than existing DDIM sampling.

\subsection{Image inpainting}

\begin{figure*}
    \centering
    \begin{subfigure}[b]{0.32\columnwidth}
                \includegraphics[width=\textwidth]{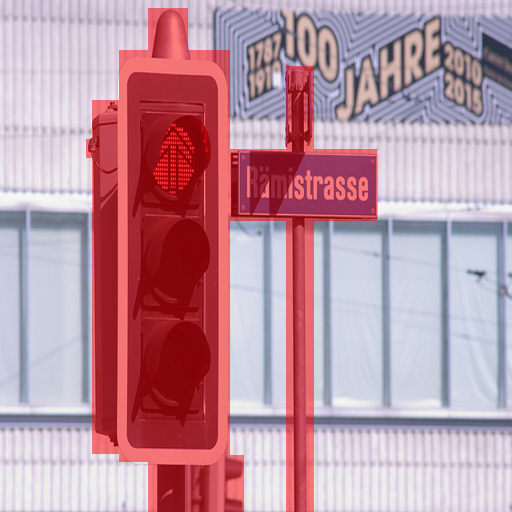}
    \end{subfigure}
    \begin{subfigure}[b]{0.32\columnwidth}
                \includegraphics[width=\textwidth]{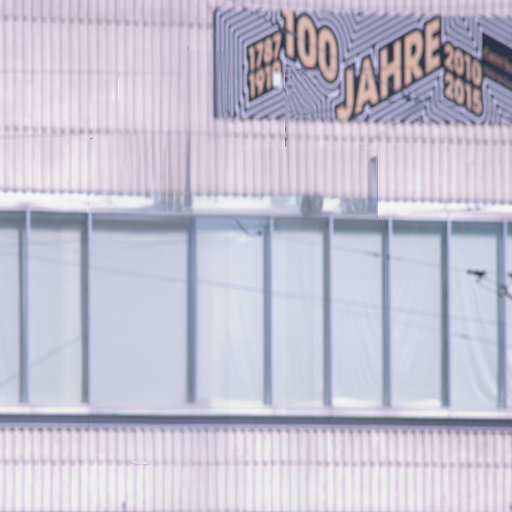}
    \end{subfigure} 
    \begin{subfigure}[b]{0.32\columnwidth}
                \includegraphics[width=\textwidth]{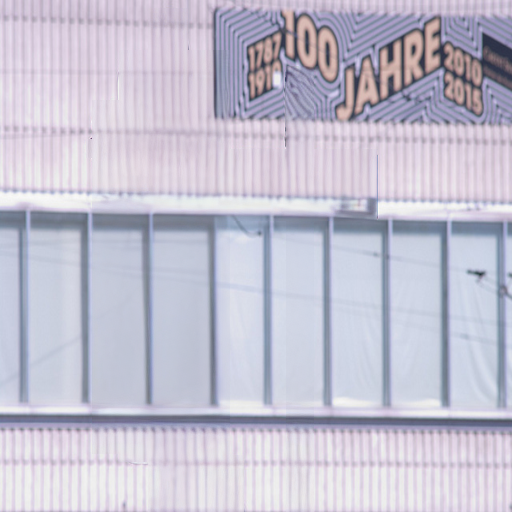}
    \end{subfigure} 
    
    \begin{subfigure}[b]{0.32\columnwidth}
                \includegraphics[width=\textwidth]{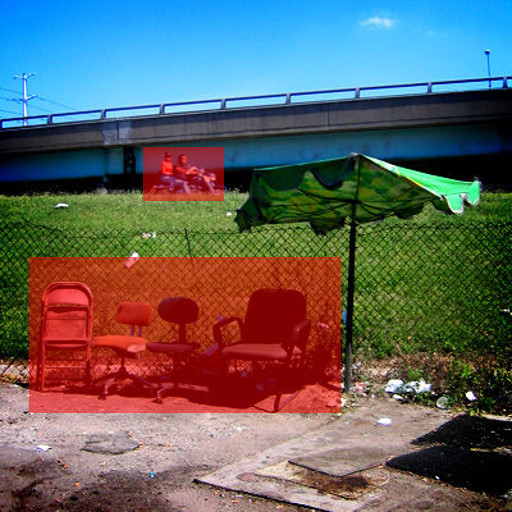}
    \end{subfigure}
    \begin{subfigure}[b]{0.32\columnwidth}
                \includegraphics[width=\textwidth]{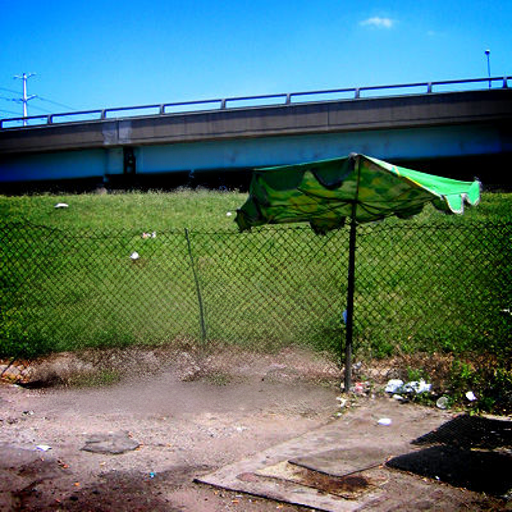}
    \end{subfigure} 
    \begin{subfigure}[b]{0.32\columnwidth}
                \includegraphics[width=\textwidth]{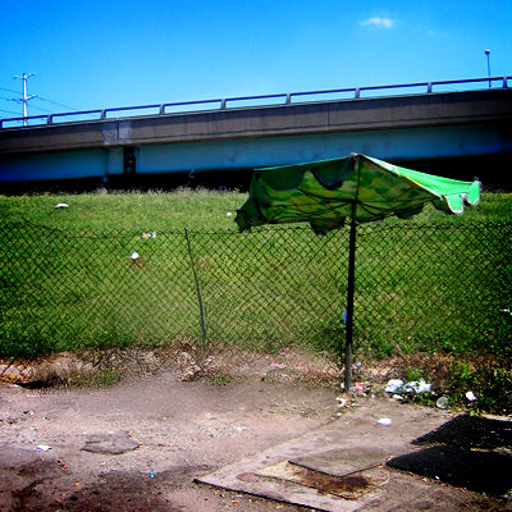}
    \end{subfigure} 
    
    \begin{subfigure}[b]{0.32\columnwidth}
                \includegraphics[width=\textwidth]{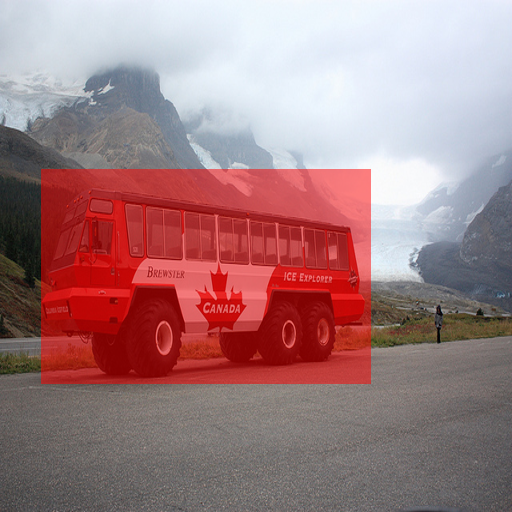}
                \caption{Source Image}
    \end{subfigure}
    \begin{subfigure}[b]{0.32\columnwidth}
                \includegraphics[width=\textwidth]{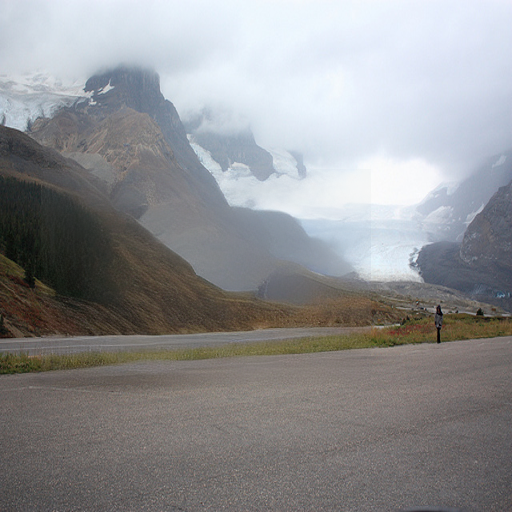}
                \caption{DDIM}
    \end{subfigure} 
    \begin{subfigure}[b]{0.32\columnwidth}
                \includegraphics[width=\textwidth]{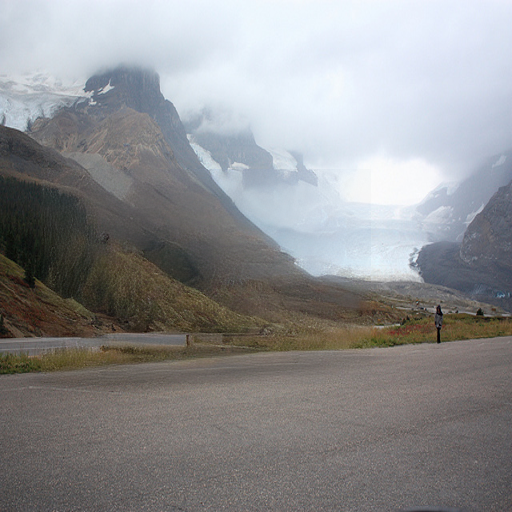}
                \caption{Ours}
    \end{subfigure} 
    
    \caption{LDM-Image Inpainting result of the source image (a). For comparison at the same latency, we compare DDIM 8 step (b) and Ours(DDIM 15 step interval 2) (c). As shown in the red box, Ours synthesizes the masked region reflecting the surrounding context more effectively than DDIM.}
    \label{fig:GRID-Inpaint}
\end{figure*}

In Fig.~\ref{fig:GRID-Inpaint}, we evaluate the performance of our method in Image Inpainting compared to DDIM sampling within the same latency budget, 1.16s. The image inpainting process generates content for the region corresponding to the mask image from the source image. Compared to DDIM (b), our method (c) generates higher-quality images and can effectively generate more content by considering the surrounding context. This suggests that our approach tends to recover more parts of the image efficiently when applying image inpainting with a smaller number of steps compared to DDIM.
\subsection{Image-to-video}

Furthermore, we assess the applicability of our method in the image-to-video model from Stability AI \footnote{https://huggingface.co/stabilityai/stable-video-diffusion-img2vid}. Our approach can be safely applied to the video diffusion model and achieve an acceleration of 1.95x without quality degradation. Please refer to the attached \texttt{video.mp4} file for the results.

\section{Potential Ethical Consideration}
\label{sec:ethical}
Because FRDiff relies solely on a pretrained diffusion model without any additional data or modifications, we believe it does not introduce any additional issues beyond the ethical concerns inherent in the model itself. Moreover, by abstaining from further data augmentation or model alterations, FRDiff maintains its integrity as a tool while mitigating potential risks associated with unintended consequences or biases introduced through additional modifications, while training based distillation methods \cite{salimans2022progressive,sdxlturbo,song2023consistency,lcm} does not.

\end{document}